\documentclass[11pt]{article}

\usepackage[preprint]{acl}
\usepackage{xspace}
\usepackage{colortbl}
\usepackage{times}
\usepackage{latexsym}
\usepackage{booktabs}
\usepackage{amsmath}
\usepackage{longtable}
\usepackage{tabularx}
\usepackage{array}
\usepackage{booktabs}
\usepackage[T1]{fontenc}
\usepackage{xcolor}
\usepackage[utf8]{inputenc}
\usepackage{array}
\usepackage{multirow}
\newcolumntype{G}{>{\centering\arraybackslash}p{1.6cm}}

\usepackage{microtype}
\newcommand{\cameraready}[1]{\xspace}
\usepackage{inconsolata}
\newcommand{\ci}[1]{{\scriptsize\color{gray}#1}}
\newcommand{\na}{\textendash}
\usepackage{amsmath}
\newcommand{\camerareadytext}[1]{\xspace}
\usepackage{array}
\newcolumntype{G}{>{\centering\arraybackslash}p{1.45cm}}
\usepackage{pgfplots}
\pgfplotsset{compat=1.18}
\usepackage{sansmath} 
\tikzset{every mark/.append style={scale=1.1}}
\usepackage{graphicx}

\title{VISTA: Steering LLMs with Latent User Values Inferred from a Few Survey Answers}

\author{Yinuo Xu${}^{1}$, Jadie Adams${}^{2}$, Brian Hu${}^{2}$, Shivani Kumar${}^{1}$, David Jurgens${}^{1}$, and Arslan Basharat${}^{2}$ \\
${}^{1}$University of Michigan, ${}^{2}$Kitware Inc. \\
\texttt{\{yinuoxu, kshivan, jurgens\}@umich.edu}, \\\texttt{\{jadie.adams, brian.hu, arslan.basharat\}@kitware.com}}

\usepackage{amsfonts}
\usepackage{tabularx}
\usepackage{tabularx}
\usepackage{booktabs}
\usepackage{array}
\usepackage{pifont}
\usepackage{ragged2e}
\renewcommand{\arraystretch}{1.2}

\newcolumntype{Y}{>{\RaggedRight\arraybackslash}X}
\begin{document}
\maketitle
\begin{abstract}

 Personalizing a language model requires knowing what a person values, which stays largely consistent
  across tasks. Yet most methods either learn task-specific preferences or place the user's profile in the
  prompt, where models struggle to act on it. We introduce VISTA (Value-Informed Steering and Targeted
  Acquisition), which infers a latent profile from users' (full or partial) survey answers, and maps the profile through a hypernetwork to a user-specific steering intervention in frozen LLM to select answer choices or generate free-text responses. Given the
  same evidence, VISTA matches or beats in-context profiles on synthetic preferences (Food2D), the World
  Values Survey (WVS), and PrimeX, 
  without adding prompt tokens. To reduce the cognitive burden of completing full surveys, VISTA also chooses which questions to ask and when to stop: on WVS it stops after 6\% of the questionnaire while retaining 94\% of the full-history gain, where standard adaptive testing retains at most 42\%. The dominant steering direction is an interpretable value
  axis on which VISTA places unseen users accurately. Our results point toward pluralistic alignment that
  adapts to individuals at deployable cost.
\end{abstract}

\section{Introduction}

\begin{figure}[!t]
    \centering
    \includegraphics[width=0.5\textwidth]{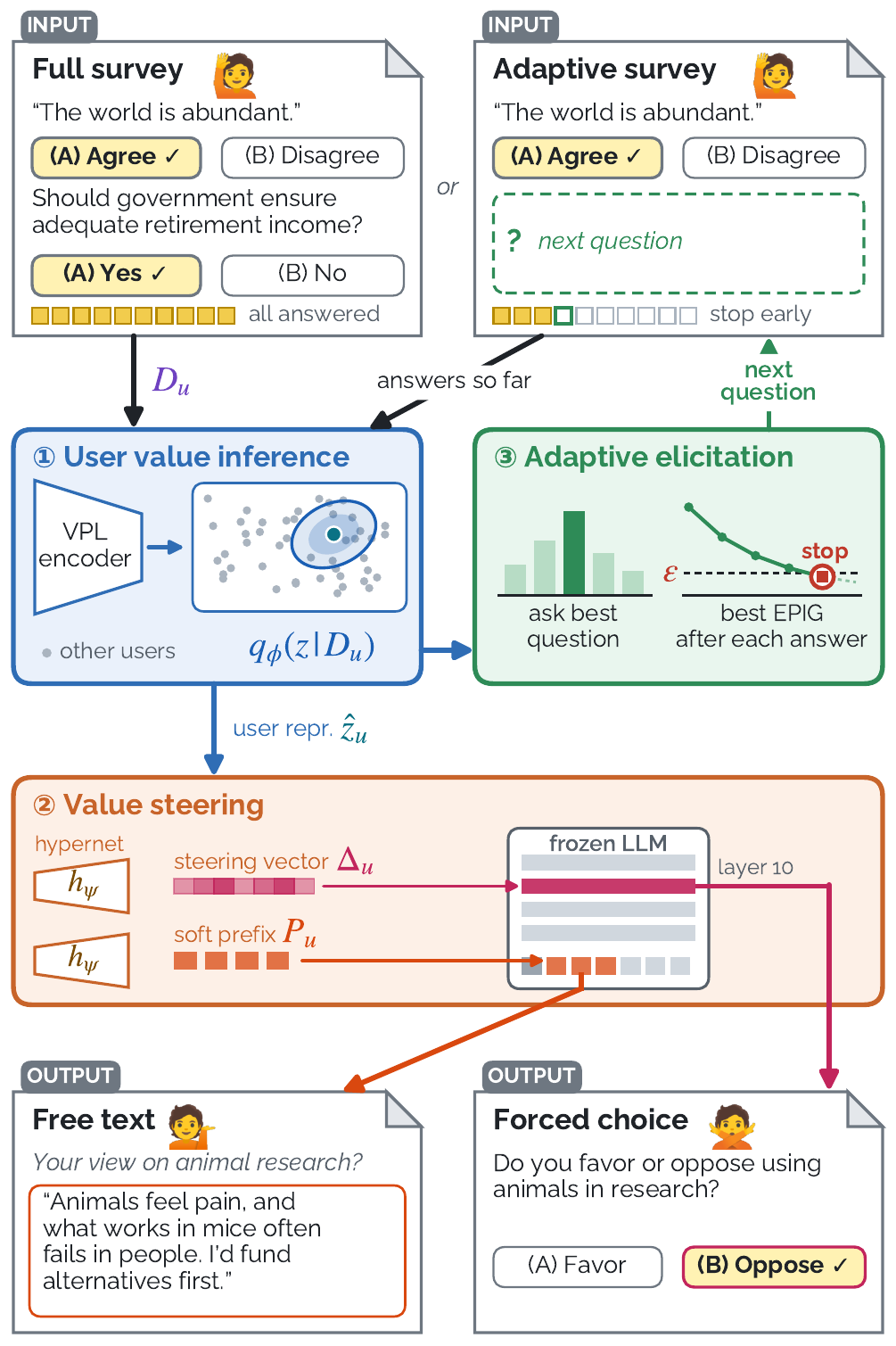}
    \caption{\textbf{Overview of VISTA.} User responses $D_u$ come from full or adaptive surveys. \textbf{\ding{172}~Value inference:} a shared encoder (grey) maps $D_u$ to a posterior $q_\phi(z \mid D_u)$ over a 32-d latent with mean $\hat z_u$. \textbf{\ding{173}~Value steering:} hypernetworks $h_\psi$ map $\hat z_u$ to a layer-10 steering vector $\Delta_u$ for forced choice or 16 soft-prefix tokens $P_u$ for free text in a frozen LLM. \textbf{\ding{174}~Adaptive elicitation} (forced choice only): select the unasked question with highest expected predictive information gain (EPIG) until its score falls below $\varepsilon$ times the initial maximum.}
    \label{fig:fig1}
\end{figure}

Aligning a language model to an individual requires knowing what that person values, not merely which of two candidate responses they happened to prefer \cite{kirk2024benefits}. Most personalization work takes the latter route, inferring a latent user representation from pairwise judgments on a task's responses \citep{poddar2024personalizingreinforcementlearninghuman,
  chen2024palpluralisticalignmentframework, bose2025lorepersonalizingllmslowrank,
  li2025prefpalettepersonalizedpreferencemodeling}, so the signal is tied to the
  task it was collected on. Values are organized differently: their structure is largely shared across people and cultures, and individuals differ in the priority they give each dimension \citep{SCHWARTZ19921}. Because these priorities hold across situations, a representation of them need not be rebuilt for each task.
  Demographic personas capture group tendencies but not individuals \citep{santurkar2023opinionslanguagemodelsreflect, liu-etal-2024-evaluating-large}, and placing the user's description in context helps \citep{joshi2025improvinglanguagemodelpersonas,
  li2026personabasedsimulationhumanopinion, moon2026virtualpersonaslanguagemodels} but does not guarantee the model acts on it \citep{kumar2025rulesmeantbrokenunderstanding}, while the prompt grows with every answer it carries. Additionally, long surveys impose a cognitive load that
  respondents manage by satisficing, answering with less care and attention as
  the task wears on, which lowers the quality of the data
  \citep{barge2012using}. A practical system should ask few
  questions, choosing which ones matter and stopping when it has asked enough.

 We introduce \textbf{VISTA}  (Value-Informed Steering and Targeted Acquisition) (Figure \ref{fig:fig1}), a framework for aligning LLM responses with individual user values.
 VISTA infers a posterior over a user's value latent from full or partial survey answers and maps it through a hypernetwork to a user-specific steering intervention in a frozen LLM.
  It uses the same posterior to select questions and decide when to stop. VISTA needs no per-user
  weight update and adds a small, fixed cost per query, unlike best-of-$N$
  reranking or long user prompts.  Prior work addresses these pieces separately: preference models personalize by reranking or guiding sampled outputs \citep{shenfeld2025languagemodelpersonalizationreward,lee2026uncertaintyawarevariationalrewardfactorization,padurean2025inferencetimepersonalizedalignmentuser}, per-user interventions
  capture writing style\citep{zhang2025personalizedtextgenerationcontrastive, popp2026predictingsteeringvectorsadapter}
  or use point estimate from randomly chosen answers\citep{li2024steerabilitylargelanguagemodels}, and adaptive elicitation uses fixed question budgets   \citep{choudhury2026bedllmintelligentinformationgathering,wang2025adaptiveelicitationlatentinformation}, or repeat the process for every new request\citep{padurean2025inferencetimepersonalizedalignmentuser}.

  VISTA is the first to infer a
  posterior over a user's values from a few survey answers, generate an
  activation-level intervention from them, and decide which questions to ask and when to stop. The value encoder also transfers: trained on the World Values Survey, it can be reused without retraining on a different population and questionnaire (PrimeX). Our contributions are: (1) \textbf{A learned latent represents a user better
  than a written profile.} Given identical evidence, VISTA matches or beats
  placing the user's answers in the prompt on all three datasets, and on PrimeX
  it is the only system whose personalized free text yields a significant net
  gain. (2) \textbf {Personalization at constant cost and under realistic question budgets}
  Each query adds one steering vector or 16 soft tokens to a frozen model, where prompting re-reads the user's answers every time. VISTA decides which questions to ask and when to stop: on WVS it stops after
  6\% of the questionnaire while retaining 94\% of the full-history gain.
  (3) \textbf{An interpretable representation.} The dominant steering direction
  is a human-readable value axis, and steering places unseen respondents at
  their own position on it ($r = 0.73$ on WVS, $0.59$ on PrimeX).


\section{Related Work}
\label{sec:related}
\noindent\textbf{Preference and user modeling.} Preference models learn shared structures from pairwise comparisons: weighted basis rewards \cite{shenfeld2025languagemodelpersonalizationreward, bose2025lorepersonalizingllmslowrank}, Gaussian bases \cite{lee2026uncertaintyawarevariationalrewardfactorization}, prototype mixtures \cite{chen2024palpluralisticalignmentframework}, variational latents \cite{poddar2024personalizingreinforcementlearninghuman}, or sociolinguistic and value dimensions for communities \cite{li2025prefpalettepersonalizedpreferencemodeling}.
These methods rely on reranking \cite{lee2026uncertaintyawarevariationalrewardfactorization, padurean2025inferencetimepersonalizedalignmentuser} or reward-guided decoding \cite{shenfeld2025languagemodelpersonalizationreward}, requiring multiple candidates or per-step evaluation.

\noindent\textbf{Activation steering.} Steering vectors are typically shared across users \citep{cao2024personalizedsteeringlargelanguage, chen2025personavectorsmonitoringcontrolling, zhu2025steeringriskpreferenceslarge, yu2026tracingmoralfoundationslarge}, with prompting outperforming vector injection for personality traits  \cite{banayeeanzade2026psychologicalsteeringllmsevaluation}. Per-user methods derive writing-style vectors from past texts \citep{zhang2025personalizedtextgenerationcontrastive} or three samples \citep{popp2026predictingsteeringvectorsadapter}, or adjust fixed-vector strengths \citep{bo2026steerablechatbotsexploringpersonalization}. Closest to VISTA, \citet{li-etal-2024-steerability} learn point-estimate respondent embeddings mapped to prefix tokens from random answers, while \citet{tan-etal-2026-instant} generate query-specific LoRA weights from writing histories. 

\noindent\textbf{Representing a user's values.} Demographic personas poorly represent individuals \cite{liu-etal-2024-evaluating-large}, while personal rationales \citep{joshi2025improvinglanguagemodelpersonas} and narratives \citep{li2026personabasedsimulationhumanopinion} help. \citet{sorensen-etal-2025-value} compress demonstrations into interpretable but lossy value profiles, without significant gains in their free-form text generation. Even structured moral profiles can yield near-chance performance \citep{kumar2025rulesmeantbrokenunderstanding}. 

\noindent\textbf{Adaptive elicitation.} Computerized adaptive testing selects informative items and stops at a standard-error threshold \cite{xu2023developing, vanderLinden2010ElementsOA}. LLM-based methods maximize information gain about natural-language hypotheses \citep{choudhury2026bedllmintelligentinformationgathering}, held-out answers \citep{wang2025adaptiveelicitationlatentinformation}, or textual preference criteria \citep{pmlr-v306-bose26a}, under fixed budgets. VPL uses fixed queries \citep{poddar2024personalizingreinforcementlearninghuman}, PReF adapts within a fixed budget \citep{shenfeld2025languagemodelpersonalizationreward}, and UserAlign stops by rule but re-elicits per prompt \citep{padurean2025inferencetimepersonalizedalignmentuser}.

VISTA infers a posterior over individual values from a few survey answers, maps it to reusable interventions in a frozen LLM with an interpretable value axis,
and adaptively selects questions and stops by expected predictive information gain.

\section{Data}
We evaluate VISTA on three datasets: synthetic food preferences (Food2D), real-world social values (WVS), and opinions and world beliefs (PrimeX), Together, they test personalization across synthetic and real-world preferences, with PrimeX additionally evaluating transfer to a new population and survey instrument. We describe each dataset in detail next:


\noindent\textbf{Food2D} \citep{padurean2025inferencetimepersonalizedalignmentuser}
  is a synthetic dataset with 10 food questions (e.g., ``What should I eat
  tonight?''), each with 20 candidate responses annotated on two features,
  spiciness and vegetarian-ness ($\varphi$). An item is a pair of responses to
  the same question (1,900 pairs). Each simulated user has a two-dimensional
  taste vector $\theta$ and chooses between two responses noisily, in
  proportion to how well each matches their taste; this noise caps accuracy at
  66\% even for a method given the true $\theta$. We sample 2,000 training
  users answering 40 pairs each and hold out 300 unseen users. Our models see
  only the response text, never $\varphi$ or $\theta$.

\noindent\textbf{World Values Survey.}
We use US and UK respondents from WVS Wave 7 \citep{wvs7}: 5,652 people (3,956/848/848 train/val/test) answering about 130 of 226 value statements across 15 topics, including religion, politics, economics, immigration, ethics, trust, science, and security. This gives 73,288 held-out (person, item) pairs. It gives us a real testbed for personalized value inference as it captures naturally occurring preferences on socially consequential topics. We converted the WVS into binary choice-questions following \citet{jiang2025languagemodelsreasonindividualistic}. There are 226 items, each rewritten as first-person statements (e.g., "religion is important in my life" vs "religion is not important in my life"), resulting in 217 items having two sides and 9 having three. Multi-point scales are split at midpoint so each answer maps to one statement.

\noindent\textbf{PrimeX} \citep{koncelkedziorski2025primexdatasetworldviewopinion} serves as a transfer test with a new population, item bank, and domain. It has 858 US respondents, each answering about 46 of 48 binary items (30 opinions on medicine, science, and risk; 18 \emph{primals}, or abstract world beliefs), along with demographics and free-text explanations. We converted PrimeX into the same format as WVS: each item is rewritten as a pair of first-person statements, one per side. Multi-point scales are split into two sides: 4-point opinion scales at split. At midpoint, and 6-point Primals scales into agree vs. disagree.

\section{Methods}
\subsection{Inferring Latent Values}

Human values are heterogeneous and cannot be fully inferred from limited enrollment survey responses. We therefore adopt the Variational Information Bottleneck (VIB) framework \cite{alemi2017deep} to learn probabilistic latent representations that capture individual preferences and uncertainty while leveraging shared structure across a population. Our approach,VISTA-VI (Value inference component of VISTA),  builds upon Variational Preference Learning (VPL) \cite{poddar2024personalizingreinforcementlearninghuman}, which models heterogeneous preferences through user-specific latent variables in personalized reward learning. We adapt this approach to value elicitation from enrollment surveys, incorporating swap-guided regularization to mitigate encoder collapse and semi-amortized inference to improve user-specific representations.

We represent each user's values through a latent variable $z \in \mathbb{R}^{32}$ , which captures unobserved factors influencing their survey responses. We set the latent dimension to 32, chosen on a validation split from $K \in {16, 32, 64}$: it is the only size at which every search trial converged (Appendix~\ref{sec:search-K}). Given a user's observed answers, an amortized encoder estimates a diagonal Gaussian posterior, $q_\phi(z | D_u) = \mathcal{N}\big(\mu_u, \mathrm{diag}(\sigma_u^2)\big)$ where $D_u$ denotes the user's enrollment responses, $\mu_u$ represents the estimated latent preference vector, and $\sigma_u^2$ captures uncertainty in that estimate. For each observed survey item, the encoder retrieves the text embedding of the user's selected response from a statement bank. These embeddings are aggregated using learned-query attention pooling to produce a fixed-dimensional representation, which is mapped to the posterior parameters $\mu_u$ and $\sigma_u$. During training, latent variables are sampled using the reparameterization trick ($z = \mu_u + \sigma_u \odot \epsilon$ where $\epsilon \sim \mathcal{N}(0,I)$) and $z = \mu_u$ is used at evaluation.

A preference decoder assigns a scalar score $s_\theta(x,z)$ to each candidate response $x$, conditioned on the user's latent representation. $z$ is injected into the statement embedding through Feature-wise Linear Modulation (FiLM) \cite{perez2017filmvisualreasoninggeneral}. Since decoder scores statement embedding instead of looking up an item by index, it can score text it never saw in training. Choice probabilities are obtained by applying a softmax over the candidate scores. This corresponds to the Bradley–Terry choice model \citep{bradley1952rank} for binary items and the top-choice Plackett–Luce model \citep{luce1977choice} for items with multiple options. 

 \noindent\textbf{Training and refinement.} The model is trained to predict
  held-out survey responses with a KL penalty toward a standard Gaussian prior.
  To prevent posterior collapse, a known failure of VPL in which the encoder
  ignores the user, we add swap-guided regularization
  \cite{kim2026swapguidedpreferencelearningpersonalized}. Because one encoder
  shared across the population can be suboptimal for an individual, we use
  semi-amortized inference \cite{kim2018semiamortizedvariationalautoencoders}:
  the encoder's posterior mean is refined for 100 gradient steps on the user's
  observed answers, with a penalty that keeps it close to the encoder's
  estimate. Objectives and training details are in
  Appendix~\ref{app:value-model}.

\subsection{Activation Steering from Inferred Values}

A hypernetwork $h_\psi$, inspired by HyperSteer \cite{sun2025hypersteeractivationsteeringscale}, maps the inferred representation $\hat z_u$ to an intervention in a frozen Llama-3-8B-Instruct. Steering vectors are usually extracted from contrastive examples on a predefined concept \cite{panickssery2024steeringllama2contrastive}, which requires choosing concepts and building contrast sets for each. A hypernetwork instead learns a mapping from the value latent to the LLM intervention. The resulting directions carry no imposed meaning, but are still interpretable (cf. Section \label{sec:interp}). The hypernetwork is a small self-attention network (four blocks, eight heads), and it is the only component trained, as gradients reach it through the frozen model. 

\noindent\textbf{Steering for forced choice.} $h_\psi$ (2.4M parameters) outputs a single steering vector $\Delta_u$, normalized to unit norm so that only its direction varies across users, which is added to the residual stream at layer 10 at every token position. The layer is chosen by a validation search (Appendix \ref{sec:search-layer}). Training minimizes cross-entropy between the model's distribution over the answer letters and the user's own answer (Appendix \ref{app:steering}). 

\noindent\textbf{Steering for free-text generation.} $h_\psi$ (15M parameters) outputs a 16-token soft prefix inserted after the BOS token. It is trained in two stages. We first distill the value model into the prefix by maximizing the likelihood of a target response the value model selects for that user: the highest-scoring of 16 candidate responses on Food2D, and an explanation written by another training respondent who chose the option the value model predicts on PrimeX. WVS is excluded because it contains no user-written free-text responses. We then refine it with personalized DPO \cite{li2024personalizedlanguagemodelingpersonalized} on length-matched pairs of explanations that take opposing positions. A control with randomly assigned targets checks that gains come from the user representation (Appendix \ref{app:steering}).

\subsection{Adaptive Elicitation Framework}
We use the probabilistic value representation to adaptively select enrollment questions that are most informative for predicting downstream preferences. Following expected predictive information gain (EPIG) \cite{smith2023predictionorientedbayesianactivelearning}, we select questions based on their expected information about target-item responses rather than simply reducing uncertainty in the latent representation.

After each response, the amortized encoder updates the user's Gaussian posterior $q_\phi(z_u \mid D_u^{(t)})$, where $D_u^{(t)}$ denotes the responses collected after $t$ questions. We draw 64 posterior samples and use question-specific logistic regression models, fitted on training respondents, to estimate answer distributions. For each unasked question, we estimate its mutual information with the target-item responses under these samples and select the highest-scoring question. The posterior is updated through a single encoder forward pass after each new answer.

\noindent \textbf{Adaptive stopping.} 
Rather than imposing a fixed question budget, we stop when the maximum remaining EPIG score falls below a fraction $\varepsilon$ of the user's initial maximum score. This allows elicitation to terminate at different question counts depending on the information gained from each user's responses.
At stopping, we infer the final user representation from the collected responses and evaluate it using the trained steering module. We compare adaptive stopping with fixed-budget elicitation on the same users and frozen hypernetwork, measuring the tradeoff between enrollment burden and personalized prediction accuracy.

\section{Experiments}
Our research questions are: \textbf{RQ1}: Can a latent value or preference profile inferred from an individual's survey responses
  steer an LLM's choices toward that individual's judgments, both
  in-domain and on a new survey instrument, and how does it compare to existing inference and steering methods?
 \textbf{RQ2}: Can we steer the model to generate free-text user opinions on these survey questions? \textbf{RQ3}: Can the inferred profile select which questions to ask and decide when enough answers have been collected to steer the model's choices?



\subsection{Latent value inference and steering (RQ1) }
\label{sec:RQ1}
\noindent \textbf{RQ1 (a) Can we infer latent user profiles?}
 We split each respondent's items into an observed context and a held-out
  set, and report held-out accuracy against two controls that remove person
  information: \emph{population} ($\bar{z}$ over training users on Food2D, the per-item training-fold average on PRIMEX, $z=0$ on WVS) and \emph{shuffled} (another
  user's representation). Gains are paired per person with 95\% person-clustered bootstrap intervals.
  Baselines are a majority predictor and two alternative user-representation
  models: LoRe writes a user's reward as a weighted sum of shared basis
  rewards \citep{bose2025lorepersonalizingllmslowrank}, and PAL
  places the user at a convex combination of learned prototype ideal points and
  scores a response by its distance to that point \citep{chen2024palpluralisticalignmentframework}.
  VISTA-VI (Value inference component of VISTA) (Ours) has the largest user-specific accuracy lift on Food2D ($+11.4$ vs.\ PAL's $+9.3$)
  and WVS ($+8.7$ vs.\ $+6.6$), and ties PAL on PrimeX, where the person effect
  is small for every model (Table~\ref{tab:inference-gain}). The gain also transfers across question types: given
  only the 18 abstract Primals items, VPL still lifts opinion prediction by
  $+2.0$ points, tied with PAL and double LoRe's. We
  build on VPL because question selection and stopping both need a posterior,
  which PAL's point estimate does not provide; PAL is carried forward as an
  ablation.

\noindent \textbf{RQ1 (b) Can we steer for answer choice selection from inferred values?} We evaluate VISTA with other baselines across Food2d, WVS, and PrimeX. On PrimeX, we do transfer learning from WVS:  A frozen VPL encoder, trained on WVS (UK and US),
  maps the user's Primals answers to a 32-dimensional value embedding $z$ (the posterior
  mean). A hypernetwork maps $z$ to a steering vector added to the residual
  stream of Llama-3-8B-Instruct at layer 10; it is trained on the training folds' PrimeX
  answers. At test time $z$ is shrunk toward the mean training embedding,
  $\bar z + s(z - \bar z)$, with $s$ chosen on a separate calibration fold. We compare VISTA with these models: \underline{(1) Zero-shot LLM.} Llama-3-8B-Instruct answers each held-out item as a multiple-choice question over its option statements, with no information about the user. We take the highest-probability option, averaging over all cyclic orderings of the options. \underline{ (2) In-context profile (LLM)} \cite{jiang2025languagemodelsreasonindividualistic} We include all (130 on WVS) of the user's
  answers into the prompt as the statements they chose. Because raw LLM option
  probabilities are poorly calibrated, we recalibrate each item's logits with a per-item
  slope and bias fitted on the training folds. \underline{(3) Per-item base rate.} This is the no- personalization comparison, where each held-out item is answered with its modal class across users, fitted leave-one-fold-out so that no evaluation person contributes to the baseline scoring them.

\noindent \textbf{Setup and metrics.} We frame each held-out item as a forced choice, selecting the highest-probability option averaged over all cyclic orderings, and comparing it with the respondent's answer. The model conditions on the respondent's complete recorded responses and is scored on a disjoint set of their unseen items. On WVS, we split 3,956 train / 848 val / 848 test respondents. There are about 130 context answers each, with $n{=}73{,}288$ scored rows.  Hypersteer turns $z$ into a steering vector added to the residual stream of Llama-3-8B-Instruct at layer 10. On PrimeX, the encoder is frozen and transferred from WVS, mapping a respondent's Primals answers  (context $B$) to $z$, shrunk toward the mean training embedding by a factor set on a separate calibration fold. There are 858 people total, too small for a train/test split, so we instead rotate 5 disjoint cross-validation folds ($n{=}12{,}141$). Food2D has 300 users with 24 observed and 16 disjoint held-out comparisons each ($n{=}4{,}800$). The main metric is accuracy on the held-out set. We measure personalization gain in two ways: \emph{population} conditions on the mean training representation, and shuffled on another respondent's. The \emph{population gain} (steered $-$ population) is what knowing \emph{this} person adds over knowing the population; the \emph{user-specific gain} (steered $-$ shuffled) additionally controls for conditioning on somebody at all.  \emph{Contested} items have train-label entropy above $0.6$ nats (WVS, Food2D) or $0.65$ nats on the \emph{other} folds (PrimeX).  All arms within a dataset are scored on identical rows, with 95\% person-clustered bootstrap intervals. 

\noindent \textbf{Results.} \textbf{(1) VISTA's user representation matches or beats prompt-based profile steering from the same evidence} (Table \ref{tab:steering}). Given every observed answer, either to the encoder or to the prompt, VISTA ties in-context conditioning on PrimeX  ($+0.41$ points) and on contested WVS items ($+0.32$), and wins on WVS overall ($+2.61$) and by nearly ten points on Food2D. \textbf{(2) VISTA's user-specific gain is a multiple of in-context's on two of the three datasets} (Table \ref{tab:steering}). Measured against another person's representation, VISTA is twice in-context's on PrimeX and about five times on Food2D.  On WVS, the two are equivalent. \textbf{(3) VISTA has the largest net gain across minority- and majority-view rows} (Tables \ref{tab:minority-majority}, \ref{tab:minority-majority-wvs}). VISTA's net gain is the highest on both PrimeX and WVS, where in-context gets -3.5 net gain.

\begin{figure*}[!t]
    \centering
    \includegraphics[width=0.8\textwidth]{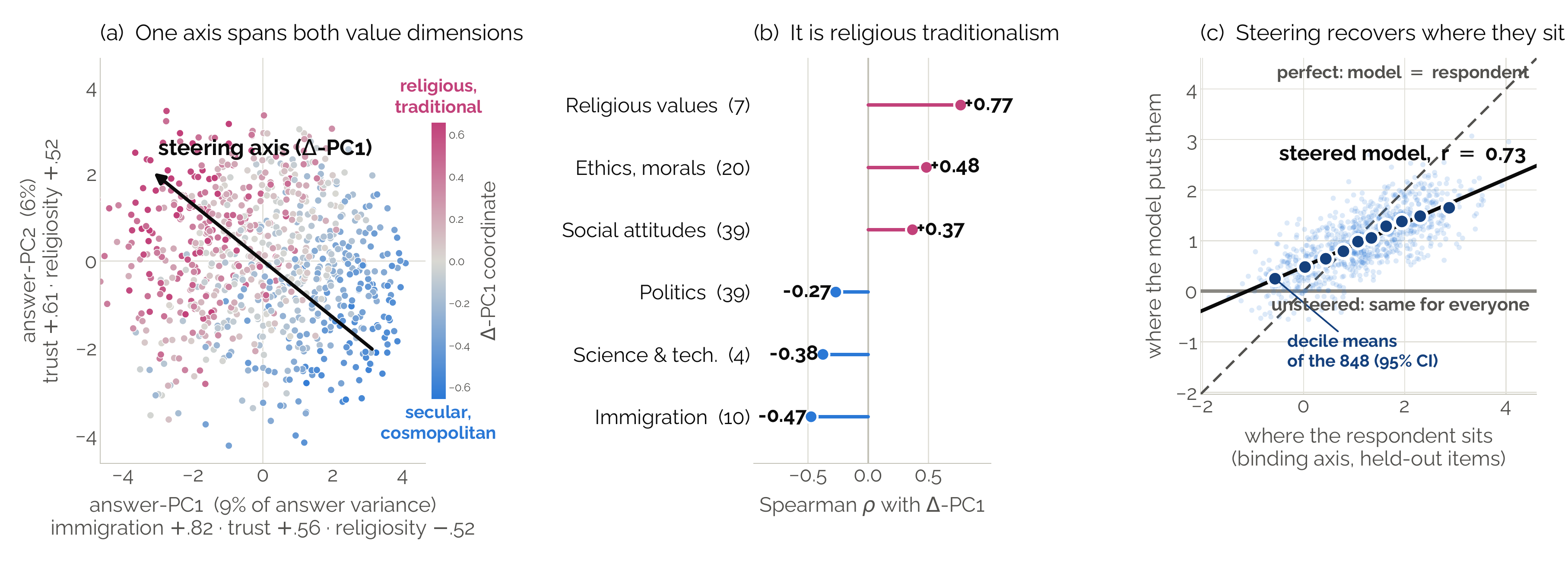}
    \caption{\textbf{The elicited value latent is a continuous, interpretable
value space, and the steered model places each held-out respondent on it.}(WVS; $\Delta=\mathrm{hypernet}(z)$ is the steering vector from a respondent's
latent.)\textbf{(a)} Respondents in the top 2 answer principal axes, colored by $\Delta$-PC1: a smooth gradient with no clusters. \textbf{(b)} Spearman correlation of $\Delta$-PC1 with each topic block's mean
answer (six largest of fourteen), where the axis spans religiosity, traditional morality, and immigration restriction, and tracks the unseen moral-foundations binding factor ($r=0.49$). \textbf{(c)} The same binding projection applied to a respondent's held-out
answers ($x$) and to the steered model's predictions on them ($y$), relative
to the unsteered model ($y=0$). }
    \label{fig:wvs-interpret}
\end{figure*}

\label{sec:interp}
\noindent\textbf{RQ1 (c) Are VISTA steering vectors interpretable?} We extract the leading principle direction ($\Delta$-PC1) of the per-respondent shift HyperSteer applies, and see what it corresponds to in the respondents' own answer space and whether the steered model places each held-out respondent correctly along it. We find that: \textbf{(1) A single interpretable steering axis spans both leading dimensions of answer variation} (Fig \ref{fig:wvs-interpret}, \ref{fig:primex-interpret}).  $\Delta$-PC1 aligns with neither leading
  answer component alone but cuts across the plane they span ($\rho = -0.72$ and
  $+0.46$ on WVS; $0.54$ and $0.33$ on PrimeX). It is religious traditionalism on
  WVS (religious values/ ethics, against immigration / science / technology) and safety versus danger on PrimeX. \textbf{(2) VISTA places unseen respondents on that axis
  where they actually sit.} The unsteered model puts everyone in the same place. The steered model recovers each held-out respondent's own position at $r = 0.73$ on WVS and $r = 0.59$ on PrimeX . UK respondents in WVS also completed a Moral Foundations
  questionnaire. We take their \emph{binding} score (the average of the loyalty,
  authority and purity foundations), find the direction in answer space that best
  tracks it among training respondents, and ask where the steered model places
  each held-out respondent along that direction. This is a demanding test
  because the value model is trained only to predict survey answers and never
  sees the moral-foundations questionnaire. The model recovers each respondent's own
  position at $r = 0.73$. VISTA has learned where this individual sits, not where people sit on average.

\subsection{Steering to Generate Free Text \camerareadytext{Based on Inferred Values} (RQ2)}
\noindent\textbf{Setup and metrics.} WVS has no user-written responses, so we evaluate on Food2D and PrimeX. Llama-3-8B-Instruct writes a free-text response to a question with no answer options, and a judge from a different model family (Qwen3.5-9B) reads it. Every system is scored against its own \emph{user-agnostic} control (no persona for prompting, the population mean $\bar z$ for VISTA), so the reported gain isolates what the user representation adds. On Food2D (23 simulated users) the prompting baselines are oracles: a persona sentence written from the user's true taste vector $\theta$, optionally with its numeric coordinates. We report how often the simulated user would prefer the generated text to the model's base response (50\% means no better). On PrimeX (686 unseen respondents) the prompting baselines receive the person's 18 Primals answers and/or demographics as text, and VISTA a soft prefix distilled from the same 18 answers. Accuracy is agreement between the side the judge reads in the text and the respondent's own survey answer. See Appendix \ref{app:rq2} for more details.

 \begin{figure}[!t]
      \centering
      \includegraphics[width=0.38\textwidth]{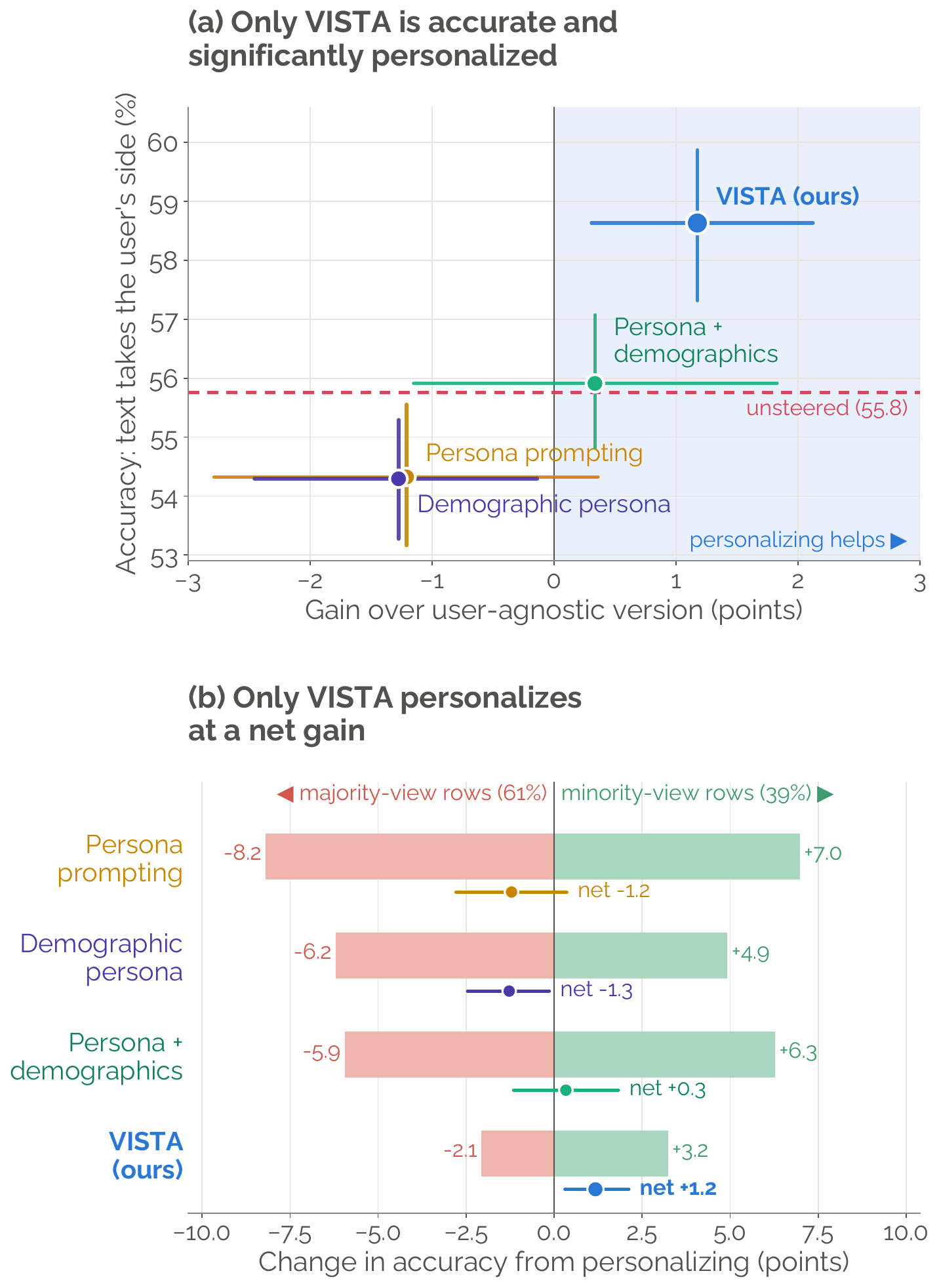}
      \caption{\textbf{Only VISTA significantly personalizes free-text generation on PrimeX.} Results cover 686 unseen users, evaluated by an LLM judge. Gain is relative accuracy to each system's user-agnostic control. \textbf{(a)} Accuracy versus personalization gain with 95\% respondent-level CIs; only VISTA's gain is significant (+1.2 points). \textbf{(b)} Gains on minority-view (39\%) and majority-view (61\%) rows.
      Unlike persona prompting, VISTA achieves a positive net gain (+1.2).}
      \label{fig:textgen_overview_1col}
  \end{figure}

  \noindent\textbf{Results.} \textbf{(1) A continuous latent carries a user better
  than a written profile.} On Food2D (Fig. \ref{fig:textgen_overview_food2d}), VISTA beats the verbal oracle by $+4.4$ accuracy points and ties the oracle that also states the
  numeric coordinates, without ever seeing the
  user's true preference vector; on PrimeX it beats unsteered and every prompting
  baseline by $+2.7$ to $+4.3$ (Fig.~\ref{fig:textgen_overview_1col}a). The
  Food2D comparison is the sharper one, because there the text baselines are
  oracles: a prose rendering of the exact user representation, and the user representation itself,
  loses to a latent inferred from answers alone. \textbf{(2) Only VISTA personalizes the content.} Only VISTA's gain over its user-agnostic control is significant on PrimeX ($+1.2$); Persona prompting gain more and lose more on majority and minority opinions,  whereas VISTA's shift is smaller and better aimed and nets positive on both groups (Fig.~\ref{fig:textgen_overview_1col}). Simply prompting produces texts that goes off topic, reciting the person's world beliefs in 77.6\% of texts (``everything happens for a reason, so\ldots''), while ours encodes the same answers but recites them in 0.9\%, close to humans.Ours also commits to a position (60.0\%) more often than prompting baselines and humans, repeats words less (3.5\%) than prompting, and is human-like in length (Fig.  \ref{fig:textgen_quality}). Tab.~\ref{tab:textgen_personalized} gives examples for two unseen users
  holding the minority view, with the side a judge reads in parentheses:
  unsteered text hedges as an assistant and persona prompting recites world
  beliefs, sometimes arguing the opposite side, while ours takes the user's side,
  often with their own reasoning.
\begin{figure*}[!tb]
    \centering
    \includegraphics[width=\textwidth]{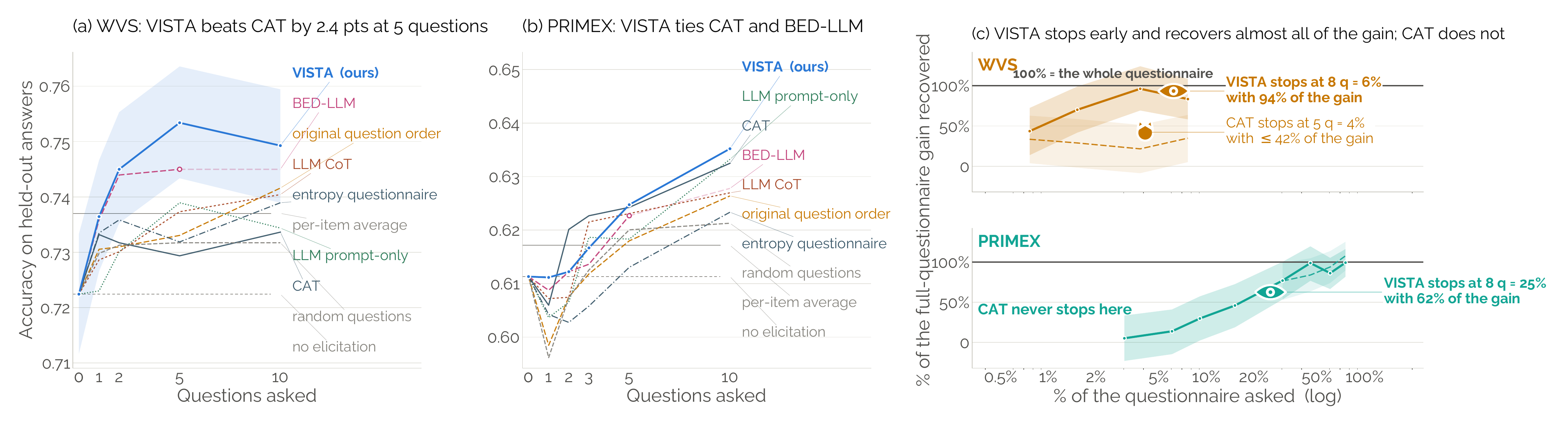}
    \caption{\textbf{Adaptive elicitation over the value latent, and when to stop.} \textbf{(a,b)} Accuracy on the held-out answer vs questions asked, for nine selection methods, across WVS (848 respondents, 6784 scored answers) and 858 respondents, 12141 answers). 95\% person-clustered intervals
  are ±1.0 points on WVS and ±0.95 on PRIMEX for every arm. (c) The same two datasets as \% of the full questionnaire's personalization gain recovered, paired over respondents against a run in which every item is observed. Bands are 95 \% respondent-clustered bootstrap intervals. Stars mark where each method's own stopping rule stops. }
    \label{fig:elicit_curve_wvs}
\end{figure*}

\subsection{Adaptive Elicitation (RQ3)}
\label{sec:elicitation}
We now evaluate a more realistic setting in which a user answers at most 10 questions instead of the full survey, and see how much gain we can achieve if we stop early. 

\noindent\textbf{Models for comparison.} Each of the following baselines both selects questions and predicts the held-out answers.  \underline{(1) CAT} \cite{xu2023developing} is the standard method for adaptive questionnaires: a two-parameter item-response model
  places each respondent on
  a single latent trait and selects the item with the
  highest Fisher
  information. CAT is the only model for comparison that has its own stopping rule (discussed shortly). \underline{(2) Prompt-only}
  \citep{zhang-etal-2024-probing}: the LLM sees the
  answers so far and 15 candidate items, names the next
  question, and predicts
  the held-out answers in context. \underline{(3) CoT}
  \citep{yao2023reactsynergizingreasoningacting}:
  prompt-only
  with an explicit reasoning step before each pick. \underline{(4) BED-LLM}
  \citep{choudhury2026bedllmintelligentinformationgathering}: a pool of 1,500
  LLM-written respondent profiles is reweighted by the
  user's answers, and the
  item with the highest expected information gain is
  asked. \underline{(5) Per-item modal answer}, which uses no
  elicitation. Details are in
  Appendix~\ref{app:elicitation}.
 \noindent\textbf{Setup and metrics.} For each budget $q
  \in \{1, 2, 5, 10\}$,
  a system selects one question at a time from the
  respondent's candidate pool,
  sees the answer, and after $q$ questions predicts that
  respondent's held-out
  answers. On WVS the pool has 131 items and 8 held-out
  items are scored per
  respondent; on PrimeX the pool has 33 items and about
  14 are scored, over 5
  folds. Food2D is reported in
  Appendix~\ref{app:elicitation}. The metric is
  held-out accuracy, also expressed as a share of the
  gain from the
  respondent's full response history. Intervals are
  paired, respondent-clustered
  bootstraps over identical held-out answers.

  \noindent\textbf{Stopping.} CAT stops when its
  posterior standard error falls
  below a threshold; we use the textbook $\mathrm{SE} <
  0.30$ and also report
  its best threshold. VISTA stops when the best remaining
  acquisition score
  falls below $\varepsilon$ times the respondent's
  first-step score, with
  $\varepsilon$ fit on calibration respondents.
\paragraph {Results.}  All percentages below are shares of the gain a respondent's \emph{full}
  response history buys, so 100\% means asking the whole questionnaire. \noindent\textbf{(1) VISTA turns questions into accuracy far faster than an
  item-response model on two of three datasets, and never worse}
  (Fig.~\ref{fig:elicit_curve_wvs}, Fig. \ref{fig:elicit-food2d}). On WVS it recovers 43.5\% of that gain from
  a single question and 70.1\% from two, out of 131 askable items, while CAT
  is flat and never exceeds 40.8\% at any budget. On Food2D it reaches 85.0\% of
  the available range against CAT's 40.4\% and Prompt-Only's 25.7\%; at $q{=}0$
  no system differs from any other, so the separation
  comes from which questions are asked and not from the systems themselves.
  PrimeX is a tie whose curves cross: CAT leads at the smallest budgets (at three questions) and we lead by ten.  On Food2D,  VISTA recovers 57.7\% of the oracle's gain from one comparison,
    65.5\% from two and 91.3\% from the full twenty, against CAT's 46.5\%, 50.1\%
    and 74.1\%. \textbf{(2) Only VISTA knows when to stop, and it stops in a
  better place than CAT.} On WVS, VISTA stops after 6.2\% of the
  questionnaire  (8.1 of 131 questions ) with 93.5\% of the gain; asking all
  ten instead is worth $+0.32$ points with no significant gain. CAT stops sooner, at 6.7 questions under its textbook $\mathrm{SE} < 0.30$, but
  recovers only 31.2\%, or 42.2\% at whichever of its five thresholds does best. Compared paired, on the same respondents answering the same items, VISTA leads
  CAT's best threshold by $+1.65$ points. On PrimeX, VISTA stops
  at 25.4\% of the pool with 62.5\% of the gain, where CAT's rule cannot fire at
  all: its median posterior SE never falls below 0.61, so no standard threshold
  is reachable and it must ask essentially the whole budget (9.6--10 questions)
  to reach a statistically indistinguishable 58.3\%. Food2D is the one dataset where CAT's rule works as intended: its median posterior SE
    falls from 1.00 to 0.18 and $\mathrm{SE} < 0.30$ triggers for 85\% of units, and
    it stops earlier than we do, at 10.0 of 20 comparisons with 93.4\% of its own
    full-budget gain against our 14.7 comparisons and 95.6\%. Stopping sooner on a flatter curve is not always a better place to stop. 


\section{Conclusion}
 We presented VISTA, which infers a posterior over a user's values from a few
  survey answers, maps it through a hypernetwork to an intervention in a frozen
  LLM, and uses the same posterior to choose questions and decide when to stop.
  Across synthetic preferences, the World Values Survey and PrimeX, the learned
  latent matches or beats placing the same answers in the prompt, at a fixed
  cost per query. On WVS, eight questions recover 94\% of the gain from a
  respondent's full history, and the dominant steering direction is a
  recognizable value axis on which unseen respondents are placed accurately.
  Together these results suggest that a compact value representation, elicited
  once and reused, is a practical route to aligning a model with an individual. We will release our code on github.

\section{Limitations}
\noindent \textbf{Eliciting informative questions for free-text generation}. Our current elicitation results are for forced choice only. Which questions are most informative for what a user would write, and when enough have been asked for generation, remains an open question. 

\noindent \textbf{Using uncertainty for steering.} VISTA carries two uncertainty signals that we only use partially. The first is the posterior variance over the value latent, which does enter question selection and stopping, but the steering intervention is generated from the posterior mean alone. The second is coherence, or how well a user's answers fit the value structure learned from the population, which identifies users with unusual profiles. Future work could use them to scale steering strength, or to ask more questions of users the model represents poorly. 

\noindent \textbf{Mitigating misalignment}. Personalization does not help on every item. When a user holds the majority view, steering toward their inferred values sometimes pulls the model away from an answer it already answered correctly: accuracy on majority-view rows falls by 3.7 points on PrimeX forced choice and by 2.1 points on free text, although the net effect is positive. Future work could detect when steering is likely to hurt and withhold it in those cases. 

\noindent\textbf{Surveys are not behavior.} We evaluate on survey answers and  on free-text opinions about survey topics. Value--behavior correlations are   modest and uneven across dimensions \citep{BARDI20031207}, attitudes predict     behavior best when both are measured at the same level of specificity\citep{AJZEN1977888}, and general moral values predict everyday behavior more weakly than context-specific ones \citep{VANDENBERG2022817860}. Whether a latent inferred from value questions steers a model well on concrete dilemmas remains open. A direct testbed is the UniMoral data \cite{kumar2025rulesmeantbrokenunderstanding}, which contains moral dilemmas annotated with labels for action choices, ethical principles, contributing factors, and consequences, alongside annotators' moral and cultural profiles.

\noindent \textbf{Generalization to model families}. All steering experiments use one frozen base model, Llama-3-8B-Instruct. Two of VISTA's three modules do not depend on the LLM: value inference embeds statements with a frozen sentence encoder and doesn't use a casual LM, and adaptive elicitation selects and stops using only the posterior over user representation $z$ learned from the value inference model. The only model-specific component in the hypernetwork, whose output lives in one model's residual stream and needs to be retrained for each base model, with finetuning on the injection layer and steering norm. 

\noindent\textbf{Risks}. A  model steered toward a user's values may reinforce their existing views, including harmful ones, and reduce exposure to other perspectives \citep{kirk2024benefits}. We do not test whether steering weakens the base model's safety behavior, and personalization should be bounded so that it does not apply to factual or safety-relevant content. 

\section{Acknowledgement} This work was conducted during the first author's internship at Kitware Inc. This material is based upon work supported by the Defense Advanced Research Projects Agency and the Air Force Research Laboratory, contract number(s): FA8650-23-C-7316. Any opinions, findings and conclusions, or recommendations expressed in this material are those of the author(s) and do not necessarily reflect the views of AFRL or DARPA.

\bibliography{custom}
\section{Appendix}
 \subsection{Compute and infrastructure}
  \label{app:compute}
  \textbf{Parameters.} The base model, Llama-3-8B-Instruct (8B parameters), is
  frozen throughout. The trained components are the value model
  (0.9\,M parameters) and the hypernetwork (2.4\,M parameters for the steering
  vector, 15\,M for the soft prefix).
  \textbf{Infrastructure.} Experiments ran on an internal SLURM cluster, each
  job on a single NVIDIA RTX A6000 (48\,GB) or A100 GPU.
  \textbf{Budget.} The full project, including hyperparameter search,
  baselines, failed runs and exploratory experiments not reported here, used
  approximately 1,800 GPU-hours.
 \subsection{Value-model search: latent size and training hyperparameters}
 \label{sec:search-K}

The value model is a variational encoder that maps a user's enrollment responses to a posterior over the latent $z$, so the search covers the quantities that affect how much information that posterior may carry. $K$ is the dimension of $z$. $\beta_{\text{target}}$ is the target weight on the KL term, i.e.\ the strength of the information bottleneck, reached through a warmup schedule. $\beta_\sigma$ is an additional multiplier applied to the \emph{width} half of the KL on top of $\beta$, which lets the pull of the posterior's scale toward the prior be set separately from the pull on its mean. The free-bits floor is applied to the mean half only. This guards against posterior collapse. $\eta$ is the encoder learning rate. The search also sweeps through a decoder learning-rate scale, which trains the decoder more slowly than the encoder. (Table~\ref{tab:search-vpl})

Table~\ref{tab:search-K} collapses the trials of Table~\ref{tab:search-vpl} by latent size. We see that $K=32$ is the only latent size at which every trial converged.
\begin{table}[t]
\centering
\small

\begin{tabular}{rrcrr}
\toprule
$K$ & Trials & Converged (NLL $<0.1$) & Best & Median \\
\midrule
16 & 6 & 2/6 & 0.0296 & 0.5061 \\
\textbf{32} & \textbf{3} & \textbf{3/3} & \textbf{0.0398} & \textbf{0.0455} \\
64 & 6 & 4/6 & 0.0460 & 0.0507 \\
\bottomrule
\end{tabular}
\caption{\textbf{Latent size, aggregated over the trials of Table~\ref{tab:search-vpl}.}
$K=32$ is the only latent size at which every trial converged. The single best trial is
$K=16$, but its advantage depends on two of six trials while the other four collapse; so we choose $K=32$ as the robust choice.
Trial counts are small and were not balanced by design (TPE allocates adaptively), so $K$=32 is a safe choice.}
\label{tab:search-K}
\end{table}

\begin{table}[t]
\centering
\footnotesize
\setlength{\tabcolsep}{3pt}
\begin{tabular}{rrrrrr}
\toprule
$K$ & $\beta_{\text{target}}$ & $\beta_\sigma\,(\times10^{-3})$ & Free bits & $\eta\,(\times10^{-4})$ & Val.\ NLL \\
\midrule
16 & 0.1066 & 2.6  & 0.362 & 9.7 & 0.0296 \\
16 & 0.0317 & 1.3  & 0.333 & 7.8 & 0.0360 \\
\textbf{32} & \textbf{0.2143} & \textbf{0.97} & \textbf{0.862} & \textbf{3.6} & \textbf{0.0398} \\
\textbf{32} & \textbf{0.0929} & \textbf{4.4}  & \textbf{0.190} & \textbf{8.0} & \textbf{0.0455} \\
64 & 0.1881 & 7.6  & 0.158 & 12  & 0.0460 \\
64 & 0.0819 & 0.11 & 0.595 & 8.2 & 0.0491 \\
\textbf{32} & \textbf{0.0338} & \textbf{0.76} & \textbf{0.127} & \textbf{8.6} & \textbf{0.0497} \\
64 & 0.0598 & 0.34 & 0.659 & 10  & 0.0503 \\
64 & 0.0347 & 1.1  & 0.103 & 13  & 0.0507 \\
16 & 0.0314 & 4.6  & 0.800 & 22  & 0.4859 \\
16 & 0.0748 & 1.2  & 0.477 & 15  & 0.5061 \\
16 & 0.0784 & 1.3  & 0.226 & 4.7 & 0.5070 \\
64 & 0.1062 & 2.7  & 0.642 & 11  & 0.5101 \\
16 & 0.2759 & 0.58 & 0.813 & 10  & 0.5109 \\
64 & 0.0784 & 2.8  & 0.100 & 6.0 & 0.5165 \\
\bottomrule
\end{tabular}
\caption{\textbf{Value-model search.} Optuna (TPE), 15 trials, objective: minimize
validation NLL. The study ran as distributed workers that number trials independently, so
rows are ordered by the objective rather than by worker-local trial id. Runs that fail to
converge land near $0.5$ (a collapsed posterior), not at a marginally worse fit.
$\beta_\sigma$ and $\eta$ are shown in units of $10^{-3}$ and $10^{-4}$, respectively.}
\label{tab:search-vpl}
\end{table}

\subsection{Optuna fine-tuning trials injection layer}. 
\label{sec:search-layer}
Our search space is layer $\in \{6, 10, 14, 20, 26\}$, HyperSteer variant $\in \{\texttt{no\_context}, \texttt{cross\_attn}\}$, $n_{\mathrm{blocks}} \in \{2, 4, 8\}$, $\delta_{\mathrm{norm}} \sim \mathcal{U}(0.3, 3.0)$, and $\mathrm{lr} \sim \log\mathcal{U}(3\times10^{-5}, 3\times10^{-4})$.
The objective is to maximize the validation user-specific effect (steered $-$ shuffled) (Table \ref{tab:search-steering}). We pick layer 10 to inject the hypersteer steering vector.

\begin{table*}[t]
\centering
\small
\label{tab:search-steering}
\begin{tabular}{llrrrl}
\toprule
Layer & Variant & Blocks & $\|\Delta\|$ & $\eta$ & Val.\ user-specific [95\% CI] \\
\midrule
\textbf{10} & \textbf{no\_context} & \textbf{8} & \textbf{1.235} & $\mathbf{2.4\times10^{-4}}$ & $\mathbf{+0.0513}$ $[+0.0416, +0.0607]$ \\
10 & no\_context & 4 & 1.476 & $6.2\times10^{-5}$ & $+0.0479$ $[+0.0386, +0.0575]$ \\
10 & no\_context & 4 & 1.282 & $1.9\times10^{-4}$ & $+0.0478$ $[+0.0382, +0.0575]$ \\
10 & no\_context & 2 & 0.831 & $7.2\times10^{-5}$ & $+0.0441$ $[+0.0354, +0.0531]$ \\
6 & no\_context & 8 & 0.692 & $4.1\times10^{-5}$ & $+0.0441$ $[+0.0343, +0.0534]$ \\
10 & no\_context & 2 & 0.940 & $1.8\times10^{-4}$ & $+0.0438$ $[+0.0349, +0.0522]$ \\
10 & no\_context & 8 & 0.723 & $8.5\times10^{-5}$ & $+0.0422$ $[+0.0335, +0.0507]$ \\
10 & cross\_attn & 4 & 0.960 & $7.3\times10^{-5}$ & $+0.0411$ $[+0.0324, +0.0497]$ \\
10 & cross\_attn & 4 & 0.716 & $3.7\times10^{-5}$ & $+0.0391$ $[+0.0305, +0.0479]$ \\
10 & no\_context & 8 & 0.646 & $1.5\times10^{-4}$ & $+0.0338$ $[+0.0256, +0.0414]$ \\
10 & no\_context & 8 & 0.617 & $1.5\times10^{-4}$ & $+0.0330$ $[+0.0251, +0.0408]$ \\
10 & no\_context & 8 & 1.015 & $1.5\times10^{-4}$ & $+0.0318$ $[+0.0234, +0.0397]$ \\
20 & no\_context & 2 & 0.313 & $1.2\times10^{-4}$ & $+0.0004$ $[-0.0012, +0.0019]$ \\
\bottomrule
\end{tabular}
\caption{\textbf{Steering-architecture search.} Optuna (TPE), 13 trials, scored on the
validation split; the test split was
never used for selection. Search space: layer $\in\{6,10,14,20,26\}$, variant $\in$
\{no\_context, cross\_attn\}, blocks $\in\{2,4,8\}$, $\|\Delta\| \sim \mathcal{U}(0.3,3.0)$,
$\eta \sim \log\mathcal{U}(3\!\times\!10^{-5}, 3\!\times\!10^{-4})$. The objective is the
validation user-specific effect (steered $-$ shuffled). Trials ran at reduced epochs; the
selected configuration was retrained at full length and scored once on test.}
\label{tab:search-steering}
\end{table*}

\subsection{Value-model training and refinement}. 
\label{app:value-model}

\noindent\textbf{Training and Collapse Prevention.} The model is trained to predict held-out survey responses while regularizing the inferred posterior toward a standard Gaussian prior. We additionally introduce swap-guided regularization to mitigate posterior collapse, a known limitation of VPL in which the encoder becomes insensitive to user-specific preferences and the decoder learns to predict population-level choices \cite{kim2026swapguidedpreferencelearningpersonalized}. The training objective is:
{\setlength{\abovedisplayskip}{2pt}
 \setlength{\belowdisplayskip}{2pt}
 \setlength{\abovedisplayshortskip}{0pt}
 \setlength{\belowdisplayshortskip}{2pt}
\begin{equation*}
    \mathcal{L} = \mathcal{L}_{\mathrm{choice}} + \beta\mathcal{L}_{\mathrm{KL}} + \lambda_{\mathrm{swap}}\mathcal{L}_{\mathrm{guide}}
\end{equation*}}
\noindent where $\mathcal{L}_{\mathrm{choice}}$ is the cross-entropy loss on held-out responses, and $\mathcal{L}_\mathrm{KL}$ regularizes the posterior toward the prior. 
The swap-guided term encourages sensitivity to individual preferences by constructing synthetic mirror users whose binary responses are reversed. Their latent representations are encouraged to exhibit opposite posterior means, $\mu^{(\mathrm{swap})} \approx -\mu$, while maintaining consistent log variances, discouraging encoder collapse.

\noindent\textbf{Semi-Amortized User Refinement.} Although amortized inference provides an efficient estimate of each user's latent preferences, a single encoder shared across the population may produce suboptimal representations for individual users. To address this amortization gap, we employ semi-amortized inference \cite{kim2018semiamortizedvariationalautoencoders}, combining population-level initialization with user-specific optimization. At inference time, the encoder provides an initial estimate $\mu_{u,0}$. We then refine this estimate through 100 Adam optimization steps, minimizing the prediction error on the user's observed enrollment responses while penalizing large deviations from the initial representation:
{\setlength{\abovedisplayskip}{2pt}
 \setlength{\belowdisplayskip}{2pt}
 \setlength{\abovedisplayshortskip}{0pt}
 \setlength{\belowdisplayshortskip}{2pt}
\begin{equation*}
    \mu_u^\star = \operatorname*{arg\,min}_{\mu} \Big[\mathcal{L}_{\mathrm{choice}}(D_u;\mu, \theta) + \tfrac{\lambda}{2}\lVert \mu - \mu_{u,0} \rVert^2 \Big]
\end{equation*}}
The regularization term balances user-specific evidence against the population-informed initialization, limiting overfitting when enrollment responses are sparse. The resulting refined latent $\mu_u^\star$ is used to predict the user's preferences on unseen questions. To ensure the decoder is compatible with these refined representations, we additionally employ an alternating optimization procedure inspired by expectation-maximization.
In the refinement (E) step, the decoder is held fixed while each user's latent mean is optimized for 100 steps against their observed responses. In the decoder-update (M) step, the decoder parameters are updated to improve predictions conditioned on the refined user representations. This procedure improves consistency between user-specific inference and preference prediction.

\subsection{Steering details}
\label{app:steering}
\noindent\textbf{Steering for forced choice.} 
The hypernetwork maps the inferred representation to a single steering vector $\Delta_u = h_\psi(\hat{z}u) \in \mathbb{R}^{d}$, where $d$ is the hidden width of the base model. We constrain $\Delta_u$ to a fixed $\ell_2$ norm, $\Delta_u \leftarrow \delta,\Delta_u / \lVert\Delta_u\rVert_2$ with $\delta=1$, so the hypernetwork determines only the \emph{direction} of the  intervention while its magnitude is held constant across users. At inference and during training we add $\alpha\Delta_u$ to the residual stream at the output of layer 10. We sweep across various layers of the residual stream of the LLM, and choose layer 10, which maximizes the user-specific effect (steered-shuffled)  (Appendix Sec \ref{sec:search-layer}). The steering vector is broadcasted over every non-padding token position, following the convention used by HyperSteer and AxBench \cite{wu2025axbenchsteeringllmssimple}, and $\alpha$ is a scalar strength multiplier fixed to $1$.  Because  $\lVert\Delta_u\rVert$ is fixed by construction, $\alpha$ is exactly the intervention magnitude, which makes it the natural place to modulate steering strength per user. The base LLM is frozen, but its activations are not detached: gradients propagate through the injection site back into $h\psi$, so the hypernetwork trains end-to-end through the frozen model. We train the additive steering module to predict held-out survey responses conditioned on each user's inferred values. For each item, the LLM assigns scores to the valid answer letters, which are normalized into a categorical distribution over the available options. The hypernetwork is trained using cross-entropy against the user's selected answer, making its performance directly comparable to the value inference model's held-out prediction
accuracy.

\noindent\textbf{Steering for free-text generation}. We first distill the value model's preferences into the soft-prefix hypernetwork through supervised likelihood training. For an inferred user representation $\hat{z}_u$, question $x$, and target response $y^\ast$  selected by a value-conditioned target policy, we minimize the response-token cross-entropy:
{\setlength{\abovedisplayskip}{2pt}
 \setlength{\belowdisplayskip}{2pt}
 \setlength{\abovedisplayshortskip}{0pt}
 \setlength{\belowdisplayshortskip}{2pt}
\begin{equation*}
    \mathcal{L}_{\mathrm{gen}} = -\tfrac{1}{T}\sum_{t=1}^{T} \log p_\theta\big(y^\ast_t \mid x,y^\ast_{<t},h_\psi(\hat{z}_u)\big),
\end{equation*}
}
where $T$ is the target response length. Prompt tokens are masked and the loss is averaged over response tokens.We then refine the hypernetwork using personalized direct preference optimization (P-DPO) \cite{li2024personalizedlanguagemodelingpersonalized}. Each preference pair contains two length-matched explanations taking opposing positions on the same question, reducing confounding from response length and wording. The personalized policy uses the user's latent representation, while the frozen reference policy uses the population-average latent $\bar{z}$ with the pre-DPO hypernetwork. Following P-DPO, we combine personalized and population-level preference terms ($\alpha=0.5$, $\beta=0.1$), using soft preference labels when the target policy is uncertain.

\noindent\textbf{Target policies}. Training targets depend on the available dataset annotations. For Food2D, we generate 16 diverse candidate responses per question offline and select the highest-scoring response under the value model's reward function $r(y, \hat{z}_u)$.
For PrimeX, the value model selects a preferred answer option, and the target is an explanation written by another training respondent who chose that option. We also evaluate an oracle variant using each respondent's own explanation.
Each setting includes a random-policy control with targets drawn from another respondent independently of the user's latent representation. This tests whether apparent gains reflect personalization rather than stylistic or evaluation biases. A distilled model should approach its policy's
pool ceiling and not exceed it by much.

\subsection{Steering to generate free text experiment details (RQ2)}
\label{app:rq2}
  \noindent\textbf{Setup and metrics.} We exclude WVS from this experiment because it does not contain any user written responses. Both Food2d and PrimeX share the same setup:
  Llama-3-8B-Instruct writes a free-text response to a question with no answer
  options, a Qwen3.5-9B judge reads the text. The judge is a different model family from the generator, so no system is rated
  by a relative of its own writer, and it sees only the question, its options and
  the generated text. It is validated separately on each dataset: on Food2D,
  calibrated per axis to GPT-4o on 300 labelled responses, it reproduces GPT-4o's
  candidate-vs-baseline winner 81.5\% of the time $[78.8, 84.1]$ (tied with
  Qwen3.5-35B-A3B); on PrimeX, judging side as A / B / unclear with option orders
  averaged, it labels the survey's own statements 18/18 and is right 95.4\% of
  the time on held-out respondents' real explanations when it commits to a side,
  abstaining on 47\%.

  Every system is scored against its own \emph{user-agnostic} control, meaning there is no persona for the prompting
  baselines, the training-population mean $\bar{z}$ for distilled prefixes (VISTA), so
  the reported gain isolates what the user representation adds. The datasets differ in what personalization is available and in what the judge extracts. On Food2D (23 simulated people) the baselines are oracles: a persona sentence rendered from the person's true taste vector $\theta$ (e.g., ``this
  user strongly loves bold heat and somewhat prefers plant-based dishes''), optionally with the raw unit-vector coordinates appended (``on a $-1$ to $+1$
  scale, spiciness: $+0.9$, veginess: $+0.4$''). The judge scores the generated
  text's own $\varphi = (\text{spiciness}, \text{veginess})$, and we report the
  simulated user's BTL probability of preferring that text to the question's base response, $\text{win} = \sigma(\theta \cdot (\varphi_{\text{text}} -
  \varphi_{\text{base}}))$, where 50\% means the text is no better than the model's base response. VISTA never sees $\theta$. On PrimeX (686
  unseen respondents) no system has an oracle: prompting baselines (persona, demographic persona, persona+demographics) receive the person's 18 world-belief (Primals) answers and/or their demographics as text, VISTA receives a 16-token
  soft prefix distilled from the same 18 answers, and generic writers (unsteered, population $\bar{z}$) receive the question alone. The judge reads which side of the topic the text takes, and accuracy is agreement with the respondent's own survey answer.

\subsection {Active elicitation experiment details (RQ3)}
\label{app:elicitation}
We evaluate our system in a more realistic setting, where users only answer up to 10 questions, instead of the full set of survey questions. We want to see if the inferred profile select which questions to ask and decide when to stop, so that a few answers suffice to steer the model's answer choices .

\noindent\textbf{Models for comparison}. We compare VISTA with other end to end systems, user beliefs, its own question-selection rule, and its own predictor for the held-out answers. \textbf{(1) 2PL CAT (Fisher Information)} \cite{xu2023developing} is the standard method for adaptive questionnaires, and the only baseline with a stopping rule. A two-parameter logistic item-response model places each respondent on a single latent trait $\theta$ and each item $j$ on a discrimination $a_j$ and a difficulty $b_j$. The posterior is:  $P(y_{ij}{=}1\mid\theta_i)=\sigma(a_j(\theta_i-b_j))$, and we fit  $(a,b)$ by marginal maximum likelihood. We use fisher information for selection, integrated over the posterior. The predictor is the IRT posterior predictive. CAT also has its own stopping rule: stop once the posterior standard error falls below a threshold, which we evaluate at the textbook operating point $\mathrm{SE}<0.30$ (reliability $0.91$). \textbf{(2) Prompt-Only (LLM)}  \citep{zhang-etal-2024-probing}: LLM is shown the answers so far and 15 shuffled candidate items, and names the next question, It then predicts the held-out answers itself, in context, from the answers it collected. The context answers then enter the prompt as the statements the person chose, where each target asked is a letter choice over its options. \textbf{(3) Chain of Thought (CoT, ReAct-Style)} \citep{yao2023reactsynergizingreasoningacting}. This is prompt-only with an explicit Thought to Action step before each pick. \textbf{(4) BED-LLM } \citep{choudhury2026bedllmintelligentinformationgathering}. We follow their preference-elicitation setup, in which a hypothesis is a paragraph of natural language describing the user. We then sample a pool of $K=1{,}500$ short respondent profiles from Llama-3-8B-Instruct and take the model's normalized distribution over each bank item's answer options, given a profile, as the likelihood  $p(y\mid\theta_k,x)$.  A respondent's answers reweight this pool and we select the item maximizing EIG. We use the same reweighted pool as its predictor. \textbf{(5) Per item modal answer}: we predict each item's most common answer, fitted on respondents who are not scored. 

\noindent\textbf{Setup and metrics}. Each system is run for a budget of $q$ questions. It selects one question at a time from the respondent's own candidate pool and is shown their answer before the next pick. Then the system predicts that respondent's held-out answers. We sweep $q\in{1,2,5,10}$. On WVS, 40\% of each respondent's answered items are held out and the remainder forms their candidate pool (131 items), with 8 held-out items scored per respondent. PRIMEX has a 48-item instrument with 33-item askable pool, and we report over 5 disjoint folds with about 14 scored items per person. On Food2d, a unit is (person, held-out question $q^*$, seed), and the pool is $9\times190=1{,}710$ comparisons from the \emph{other} nine questions. Each system is scored on 8 held-out comparisons from  $q^*$, which it was never allowed to ask about. This cross-question setup is different from the forced-choice experiment on Food2d in Section~\ref{sec:RQ1}, where we condition on a respondent's \emph{complete} recorded response set (since our research question there is representational). Here, the object of the study is about budget. So we require pool and scored items to come from disjoint questions. Because Food2d answers are Bradley-Terry draws, accuracy has a noise ceiling of $0.666$ over all pairs against a $0.496$ modal floor, so we also report it as a fraction of that range, as Section \ref{sec:RQ1} does. The main metric for all three datasets is accuracy on held-out. We also report gain over the budget: $\mathrm{acc}(q_{\max})-\mathrm{acc}(q_{\min})$, which measures whether a system uses the answers it collects.  All intervals are paired and respondent-clustered: every system is scored on the same held-out answers of the same people, and we bootstrap the per-answer difference 10{,}000 times over the respondents. 
 \noindent\textbf{Stopping.} The two rules differ. CAT stops when its posterior SE falls below a threshold; at $\mathrm{SE}<0.30$ that is 6.72 questions on WVS, triggers for 60\% of units at 7.36 questions on Food2d, and never triggers within budget on PrimeX (median SE $0.613$). VISTA stops when acquisition score falls below $\varepsilon$ times that respondent's own first-step score, with $\varepsilon$ fit on calibration respondents: it asks 8.1 questions and retains 93.6\% of the gain available from the complete response history, with no measurable loss against the full ten-question budget  ($+0.32$ $[-0.19,+0.85]$). CAT recovers at most 34.8\% of that gain at any budget we measured (33.5\%, 28.9\%, 21.5\%, 34.8\% at $q=1,2,5,10$), so no threshold choice can improve it.

\begin{table*}[t]
\centering
\small
\setlength{\tabcolsep}{4pt}
\begin{tabular}{ll cc cc cc cc cc}
\toprule
& & \multicolumn{2}{c}{\textbf{Person-cond.\ ($z$)}} & \multicolumn{2}{c}{\textbf{Population}} & \multicolumn{2}{c}{\textbf{Shuffled}} & \multicolumn{2}{c}{\textbf{Gain vs pop.}} & \multicolumn{2}{c}{\textbf{Gain vs shuf.}} \\
\cmidrule(lr){3-4} \cmidrule(lr){5-6} \cmidrule(lr){7-8} \cmidrule(lr){9-10} \cmidrule(lr){11-12}
\textbf{Data} & \textbf{Method} & All & Cont. & All & Cont. & All & Cont. & All & Cont. & All & Cont. \\
\midrule
\multirow{3}{*}{Food2D} & VISTA-VI & \textbf{0.625} & \textbf{0.624} & 0.497 & 0.494 & 0.511 & 0.513 & \textbf{+12.8*} & \textbf{+13.0*} & \textbf{+11.4*} & \textbf{+11.1*} \\
 & & & & & & & & \ci{[+11.0, +14.5]} & \ci{[+11.1, +14.8]} & \ci{[+9.1, +13.8]} & \ci{[+8.8, +13.5]} \\
 & PAL & 0.600 & 0.599 & 0.497 & 0.494 & 0.508 & 0.507 & +10.4* & +10.5* & +9.3* & +9.2* \\
 & & & & & & & & \ci{[+8.4, +12.2]} & \ci{[+8.5, +12.5]} & \ci{[+7.2, +11.4]} & \ci{[+7.0, +11.4]} \\
 & LoRe & 0.591 & 0.591 & 0.497 & 0.494 & 0.507 & 0.506 & +9.4* & +9.7* & +8.4* & +8.5* \\
 & & & & & & & & \ci{[+7.5, +11.4]} & \ci{[+7.7, +11.7]} & \ci{[+6.2, +10.7]} & \ci{[+6.3, +10.8]} \\
\midrule
\multirow{3}{*}{PRIMEX B} & VISTA-VI & \textbf{0.632} & \textbf{0.600} & 0.615 & 0.575 & 0.612 & 0.572 & \textbf{+1.8*} & \textbf{+2.5*} & +2.0* & +2.8* \\
 & & & & & & & & \ci{[+1.1, +2.5]} & \ci{[+1.5, +3.4]} & \ci{[+1.3, +2.7]} & \ci{[+1.8, +3.8]} \\
 & PAL & 0.631 & 0.598 & 0.615 & 0.575 & 0.611 & 0.569 & +1.6* & +2.3* & \textbf{+2.1*} & \textbf{+2.9*} \\
 & & & & & & & & \ci{[+1.0, +2.3]} & \ci{[+1.4, +3.2]} & \ci{[+1.4, +2.7]} & \ci{[+2.0, +3.7]} \\
 & LoRe & 0.625 & 0.590 & 0.615 & 0.575 & 0.615 & 0.575 & +1.0* & +1.5* & +1.0* & +1.5* \\
 & & & & & & & & \ci{[+0.4, +1.6]} & \ci{[+0.6, +2.3]} & \ci{[+0.4, +1.6]} & \ci{[+0.6, +2.3]} \\
\midrule
\multirow{3}{*}{PRIMEX AB} & VISTA-VI & 0.637 & 0.604 & 0.615 & 0.575 & 0.608 & 0.566 & +2.2* & +2.9* & +2.9* & +3.8* \\
 & & & & & & & & \ci{[+1.4, +3.0]} & \ci{[+1.8, +3.9]} & \ci{[+2.0, +3.7]} & \ci{[+2.7, +4.9]} \\
 & PAL & \textbf{0.638} & \textbf{0.606} & 0.615 & 0.575 & 0.603 & 0.559 & \textbf{+2.3*} & \textbf{+3.1*} & \textbf{+3.5*} & \textbf{+4.6*} \\
 & & & & & & & & \ci{[+1.7, +3.0]} & \ci{[+2.2, +4.0]} & \ci{[+2.8, +4.2]} & \ci{[+3.7, +5.6]} \\
 & LoRe & 0.632 & 0.597 & 0.615 & 0.575 & 0.610 & 0.569 & +1.7* & +2.2* & +2.1* & +2.8* \\
 & & & & & & & & \ci{[+1.0, +2.4]} & \ci{[+1.2, +3.1]} & \ci{[+1.4, +2.8]} & \ci{[+1.8, +3.7]} \\
\midrule
  \multirow{3}{*}{WVS7 UK+US} & VISTA-VI & \textbf{0.789} & \textbf{0.711} & 0.727 & 0.587 & 0.702 & 0.551 &
  \textbf{+6.2*} & \textbf{+12.3*} & \textbf{+8.7*} & \textbf{+16.0*} \\
   & & & & & & & & \ci{[+5.9, +6.6]} & \ci{[+11.7, +13.0]} & \ci{[+8.3, +9.2]} & \ci{[+15.1, +16.9]} \\
   & PAL & 0.780 & 0.694 & 0.728 & 0.588 & 0.714 & 0.586 & +5.2* & +10.7* & +6.6* & +10.8* \\
   & & & & & & & & \ci{[+4.8, +5.6]} & \ci{[+9.9, +11.4]} & \ci{[+6.1, +7.1]} & \ci{[+10.0, +11.7]} \\
   & LoRe & 0.759 & 0.667 & 0.702 & 0.549 & 0.696 & 0.570 & +5.8* & +11.8* & +6.3* & +9.7* \\
   & & & & & & & & \ci{[+5.4, +6.2]} & \ci{[+11.1, +12.5]} & \ci{[+5.8, +6.8]} & \ci{[+8.9, +10.6]} \\
\bottomrule
\end{tabular}
\caption{\textbf{Value inference only: how much of each model's accuracy is
  about the person.} Held-out answer accuracy for the three value models, two
  controls that strip person information, and the paired gains over each; bold
  marks the best person-conditioned accuracy and gain per dataset.
  \emph{Population} knows no person ($\bar{z}$ over training users on Food2D, the
  per-item training-fold average on PRIMEX, $z=0$ on WVS). \emph{Shuffled}
  deranges the test users' latents so nobody gets their own, which also removes
  the marginal fit of the latent distribution, making gain vs shuffled the
  user-specific lift and usually the stricter quantity. Gains are accuracy
  points, paired per person on identical rows, with 95\% person-clustered
  bootstrap intervals (4{,}000 resamples); ${}^{*}$ marks an interval clear of
  zero. No steering arms appear here. \textbf{Contested} uses each dataset's own
  threshold --- train-label entropy $>0.60$ nats on Food2D ($4{,}380/4{,}800$)
  and WVS ($33{,}932/73{,}288$), leave-fold-out answer entropy $>0.65$ on PRIMEX
  ($8{,}615/12{,}141$).So it is exact within a dataset but not comparable
  across them. Two readings: (i) gains are larger on contested items, as
  expected, since where the population already agrees there is nothing for a
  person model to add; (ii) VISTA-VI has the largest user-specific lift on Food2D
  ($+11.4$ vs PAL's $+9.3$) and WVS ($+8.7$ vs $+6.6$), PAL on PRIMEX AB ($+3.5$
  vs $+2.9$). Every cell is paired: WVS's PAL and LoRe population arms were added
  by re-scoring the published checkpoints, which reproduce their published
  accuracies to $2\times10^{-4}$. One exception to the usual ordering: LoRe's WVS
  population arm ($0.549$ contested) falls \emph{below} its shuffled arm
  ($0.570$), so there population is the weaker null. WVS's contested mask was
  built separately for VISTA-VI ($33{,}932$ rows) and PAL/LoRe ($33{,}632$); the All
  column is the same $73{,}288$ rows for every arm.}
\label{tab:inference-gain}
\end{table*}


\begin{figure*}[!tb]
    \centering
    \includegraphics[width=0.8\textwidth]{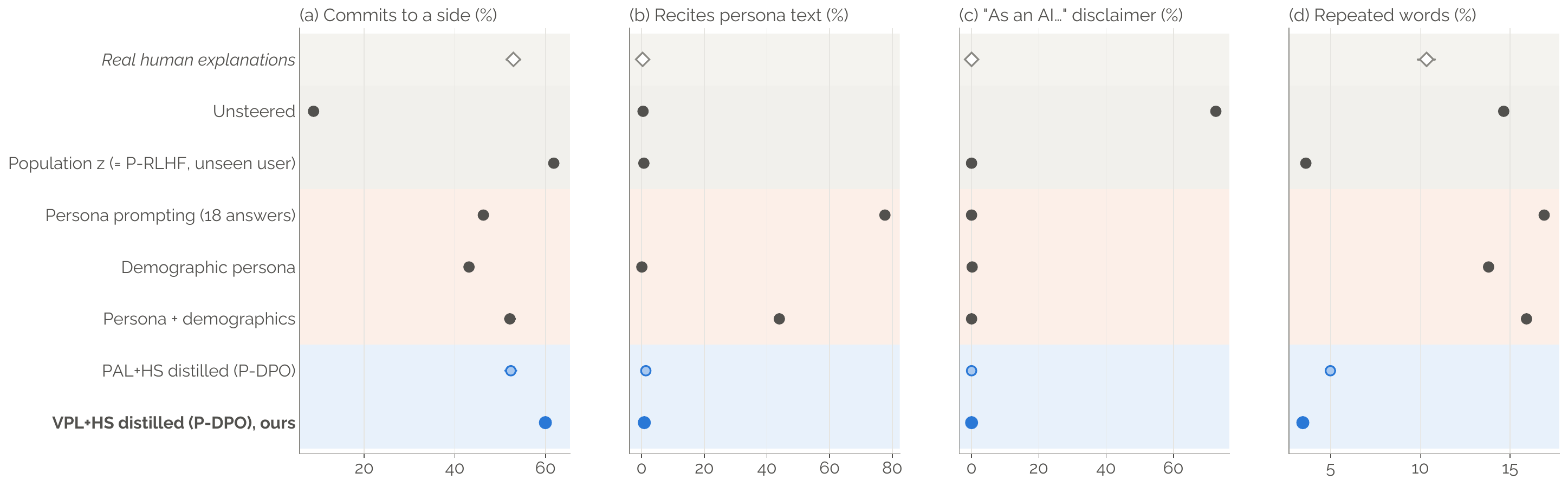}
    \caption{Personalized text generation on PrimeX.}
    \label{fig:textgen_quality}
\end{figure*}

\begin{table*}[t]
\centering\footnotesize
\setlength{\tabcolsep}{3pt}
\renewcommand{\ci}[2]{\newline{\scriptsize[#1,\,#2]}}

\begin{tabular}{@{}ll ccc GG ccc GG@{}}
\toprule
& & \multicolumn{5}{c}{\textbf{All pairs}} & \multicolumn{5}{c}{\textbf{Contested items}} \\
\cmidrule(lr){3-7}\cmidrule(lr){8-12}
& & \multicolumn{3}{c}{Accuracy (\%)} & \multicolumn{2}{c}{Gain (pts)}
& \multicolumn{3}{c}{Accuracy (\%)} & \multicolumn{2}{c}{Gain (pts)} \\
\cmidrule(lr){3-5}\cmidrule(lr){6-7}\cmidrule(lr){8-10}\cmidrule(lr){11-12}
Data & System & Steer. & Pop. & Shuf. & vs pop. & vs shuf.
              & Steer. & Pop. & Shuf. & vs pop. & vs shuf. \\
\midrule
\multirow{2}{*}{\shortstack[l]{\textbf{WVS7 UK+US}\\848 resp.\\$n=73{,}288$\\cont.\ $33{,}932$}}
 & \textbf{VISTA} & \textbf{64.87} & 63.31 & 62.58 & $+1.56$\ci{+1.21}{+1.91} & \textbf{+2.28}\ci{+1.84}{+2.72}
 & \textbf{55.48} & 51.93 & 51.89 & $+3.55$\ci{+3.09}{+4.03} & $+3.60$\ci{+3.03}{+4.16} \\
 & In-context profile & 62.25 & \na & 60.44 & \na & $+1.81$\ci{+1.45}{+2.20}
 & 55.16 & \na & 50.56 & \na & \textbf{+4.60}\ci{+4.08}{+5.11} \\
\midrule
\multirow{2}{*}{\shortstack[l]{\textbf{PrimeX} ctx $B$\\858 people\\5 folds\\$n=12{,}141$\\cont.\ $8{,}615$}}
 & \textbf{VISTA} & \textbf{63.26} & 60.85 & 58.00 & $+2.41$\ci{+1.70}{+3.10} & \textbf{+5.26}\ci{+4.15}{+6.35}
 & \textbf{59.99} & 56.61 & 53.80 & $+3.38$\ci{+2.36}{+4.34} & \textbf{+6.19}\ci{+4.79}{+7.56} \\
 & In-context profile & 62.85 & \na & 60.21 & \na & $+2.64$\ci{+1.88}{+3.39}
 & 59.36 & \na & 55.68 & \na & $+3.68$\ci{+2.63}{+4.75} \\
\midrule
\multirow{2}{*}{\shortstack[l]{\textbf{Food2D}\\300 sim.\ users\\$n=4{,}800$\\cont.\ $4{,}380$}}
 & \textbf{VISTA} & \textbf{61.44} & 50.40 & 49.35 & $+11.04$\ci{+9.06}{+12.85} & \textbf{+12.08}\ci{+9.96}{+14.25}
 & \textbf{61.23} & 50.82 & 49.50 & $+10.41$\ci{+8.41}{+12.23} & \textbf{+11.74}\ci{+9.56}{+13.97} \\
 & In-context profile & 51.86 & \na & 49.35 & \na & $+2.51$\ci{+1.21}{+3.95}
 & 51.75 & \na & 49.81 & \na & $+1.94$\ci{+0.60}{+3.43} \\
\bottomrule
\end{tabular}
\caption{Steering accuracy and personalization gain, three datasets. All arms
within a dataset are scored on identical rows; 95\% CIs from a person-clustered
bootstrap. Bold marks the better of the two arms wherever both are defined.
\emph{Population} conditions on the mean training representation,
\emph{shuffled} on another respondent's. \textbf{VISTA} is the hypernetwork
conditioned on VPL's $z$. The in-context baseline puts the person's own
answers in the prompt instead; it has no population condition---its arms are
the user's own profile, another user's, and none---and it receives the same
evidence the encoder does, every observed answer, so the comparison is about
how the person is represented and not about how much is known. Contested items
have train-label entropy $>0.6$ nats (WVS, Food2D); PrimeX contested items
exceed $0.65$ nats on the \emph{other} folds, so the split never sees the
scored fold. \textbf{Food2D} is the simulated UserAlign domain: 300 held-out
users $\times$ 16 held-out pairs, and accuracy is the steered model's
forced-choice letter against the simulated user's own answer, so the metric is
the same quantity as on the two real datasets. Its VISTA arm is the $z$-prefix
hypernetwork ($k=16$ soft tokens); its in-context arm reads the same $z$ export,
so the two rows are scored on byte-identical pairs, and paired on them VISTA
leads by $+9.57$ points $[+7.83,+11.33]$. Food2D items are near-balanced by
construction, so $91\%$ of rows ($4{,}380$ of $4{,}800$) pass the $0.6$-nat rule
and the contested subset is nearly the whole sample---which is why its two
halves agree, unlike WVS, where the contested half is the part of the bank that
separates methods at all.}
\label{tab:steering}
\end{table*}

\begin{table*}[t]
\centering
\small
\begin{tabular}{lcccc}
\toprule
& \multicolumn{2}{c}{Minority-view rows (38\%)} & Majority-view rows (62\%) & \\
\cmidrule(lr){2-3}\cmidrule(lr){4-4}
System & Accuracy & Gain & Change & Net gain \\
\midrule
In-context LLM   & 18.1 & +15.4 & $-$7.8 & +1.1 \\
Text-pooled + HS & 15.6 & +10.6 & $-$5.0 & +1.0 \\
PAL + HS         & 19.1 & +13.0 & $-$6.6 & +1.0 \\
\textbf{VPL + HS (ours)} & \textbf{24.4} & +11.2 & \textbf{$-$3.7} & \textbf{+2.0} \\
\bottomrule
\end{tabular}
\caption{Where the personalization gain comes from (PrimeX, setting B: 18
Primals answers as context; 858 unseen users, 12,141 rows). A row is
\emph{minority-view} when the user's answer differs from the most common answer
among users in the other folds. Gain and change are in accuracy points over
the same system's user-agnostic version (population representation; no profile
for the in-context LLM). Net gain is the share-weighted sum of the two columns.
All gains and changes have 95\% person-bootstrap CIs excluding 0.}
\label{tab:minority-majority}
\end{table*}

  \begin{table*}[t]
  \centering
  \small
  \begin{tabular}{lcccc}
  \toprule
  & \multicolumn{2}{c}{Minority-view rows (27\%)} & Majority-view rows (73\%) & \\
  \cmidrule(lr){2-3}\cmidrule(lr){4-4}
  System & Accuracy & Gain & Change & Net gain \\
  \midrule
  Text-pooled + HS & 32.1 & $+$2.8 & $-$0.2$^{\dagger}$ & $+$0.6 \\
  PAL + HS         & 43.0 & $+$9.4 & $-$1.5 & $+$1.5 \\
  \textbf{VPL + HS (ours)} & 40.7 & $+$6.4 & \textbf{$-$0.3}$^{\dagger}$ & \textbf{$+$1.6} \\
  \midrule
  In-context LLM$^{\ddagger}$ & 16.3 & $+$2.0 & $-$5.6 & $-$3.5 \\
  \bottomrule
  \end{tabular}
  \caption{Where the personalization gain comes from (WVS7 UK+US; 848 unseen
  respondents, 73,288 rows). A row is \emph{minority-view} when the respondent's
  answer differs from the most common answer to that item among training
  respondents. Gain and change are accuracy points over the same system's
  user-agnostic version (the population representation). Net gain is the
  share-weighted sum. 95\% person-bootstrap CIs; $^{\dagger}$ marks the two cells
  whose CI includes zero, all others exclude it. $^{\ddagger}$The in-context LLM
  has no population condition, so it is referenced against another respondent's
  profile rather than a user-agnostic one and is not directly comparable; its
  no-profile reference gains on both subsets ($+$8.9 / $+$8.0) because a bare
  prompt lacks the generic shift the steered population condition already has.}
  \label{tab:minority-majority-wvs}
  \end{table*}

\definecolor{sideyes}{HTML}{3F9B6B}
\definecolor{sideno}{HTML}{D0574A}
\definecolor{sidenone}{HTML}{8A8984}
\definecolor{oursbg}{HTML}{E8F1FB}
\definecolor{qbg}{HTML}{F1F0EC}
\begin{table*}[t]
\centering
\small
\renewcommand{\arraystretch}{1.15}
\begin{tabular}{@{}>{\raggedright\arraybackslash}p{0.19\textwidth}>{\raggedright\arraybackslash}p{0.77\textwidth}@{}}
\toprule
\rowcolor{qbg}\multicolumn{2}{@{}p{0.97\textwidth}@{}}{\textbf{Q: What's your view on the federal government's role in providing retirement income?}} \\
\rowcolor{qbg}\multicolumn{2}{@{}p{0.97\textwidth}@{}}{\textit{User's answer: Adequate retirement income is not the federal government's responsibility to provide (minority view)}} \\
\midrule
Real explanation & It is something that is more the individual responsibility with a little help from government but people need to be responsible for themselves \\
Unsteered \newline\textcolor{sidenone}{(no side)} & A timely and thought-provoking question! The role of the federal government in providing retirement income is a complex and debated topic. Here's my neutral AI perspective: Pros of government involvement: \ldots \\
Persona prompting \newline\textcolor{sideno}{(other side)} & I think the federal government should take a more active role in providing retirement income. It seems to me that what happens in life is not always meant to happen, and it's only natural \ldots \\
\rowcolor{oursbg}\textbf{Ours} \newline\textcolor{sideyes}{(user's side)} & They have no business being involved in retirement income. \\
\midrule
\rowcolor{qbg}\multicolumn{2}{@{}p{0.97\textwidth}@{}}{\textbf{Q: What's your view on modern medical treatments and whether they are worth their costs?}} \\
\rowcolor{qbg}\multicolumn{2}{@{}p{0.97\textwidth}@{}}{\textit{User's answer: Medical treatments these days often create as many problems as they solve (minority view)}} \\
\midrule
Real explanation & I feel like in most cases people who are very sick can't afford to get better, so therefore, if they try, they have a great cost of them which causes stress and \ldots \\
Unsteered \newline\textcolor{sideno}{(other side)} & A complex and multifaceted topic! As a neutral AI, I'll provide an objective analysis of modern medical treatments and their costs. Advances in Medicine Modern medicine has made tremendous progress \ldots \\
Persona prompting \newline\textcolor{sideno}{(other side)} & I tend to see the world as a pretty safe place, and I believe that modern medical treatments are a testament to humanity's capacity for innovation and progress. While some things may not be \ldots \\
\rowcolor{oursbg}\textbf{Ours} \newline\textcolor{sideyes}{(user's side)} & They cost an arm and a leg. They are not regulated well enough to ensure they work. And most times the side effects outweigh any potential benefits. \\
\bottomrule
\end{tabular}
\caption{\textbf{Illustrative VISTA generated texts on PrimeX} for two unseen users holding the minority view; parentheses give the side a judge reads in each text. Drawn at random from rows where our text takes the user's side only under their own profile (8.1\% of minority-view rows, vs.\ 5.7\% flipping away).}
\label{tab:textgen_personalized}
\end{table*}

\begin{figure}[!t]
    \centering
    \includegraphics[width=0.6\textwidth]{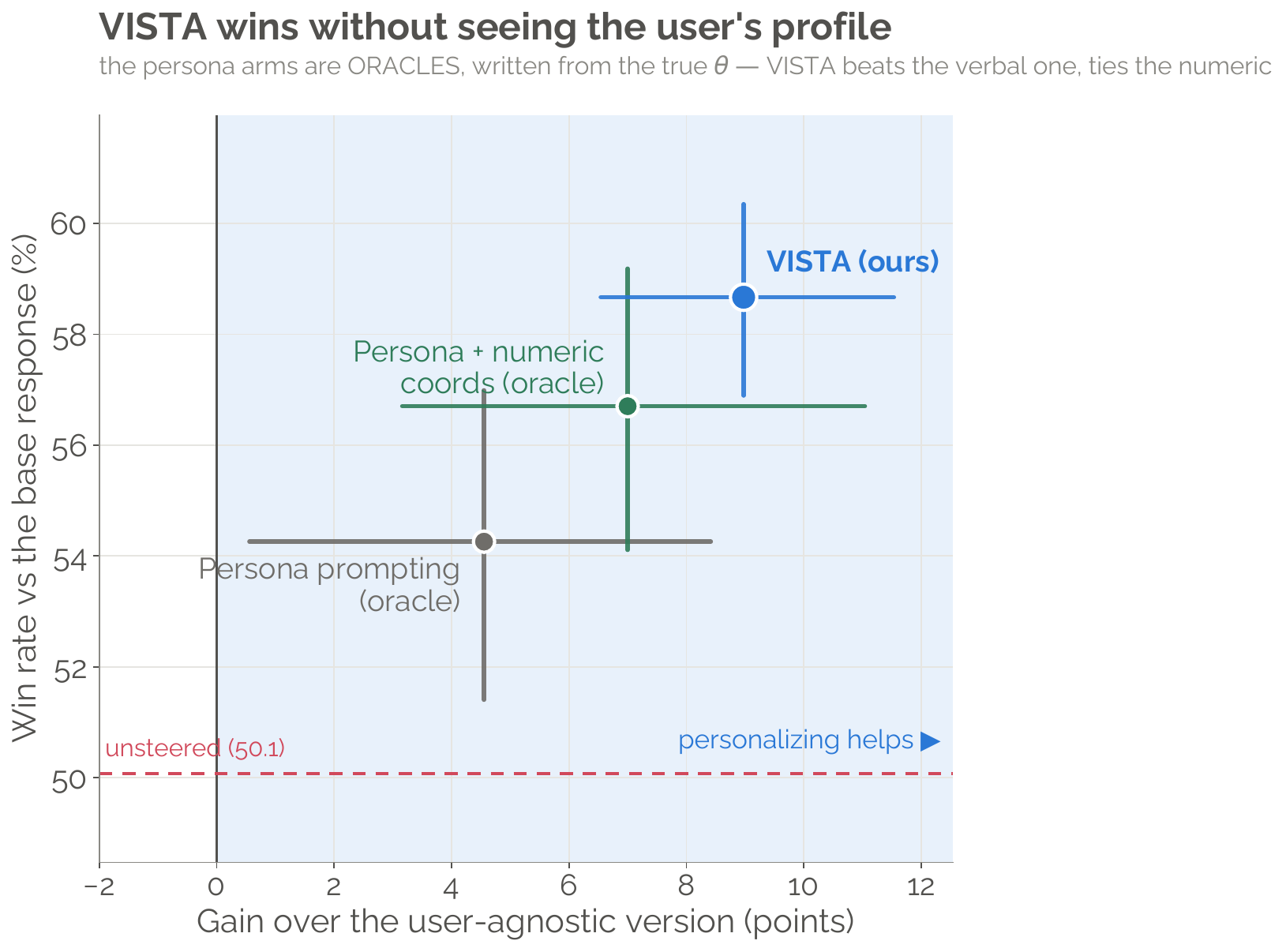}
    \caption{Personalized text generation on Food2d}
    \label{fig:textgen_overview_food2d}
\end{figure}

\begin{figure}[!t]
    \centering
    \includegraphics[width=0.7\textwidth]{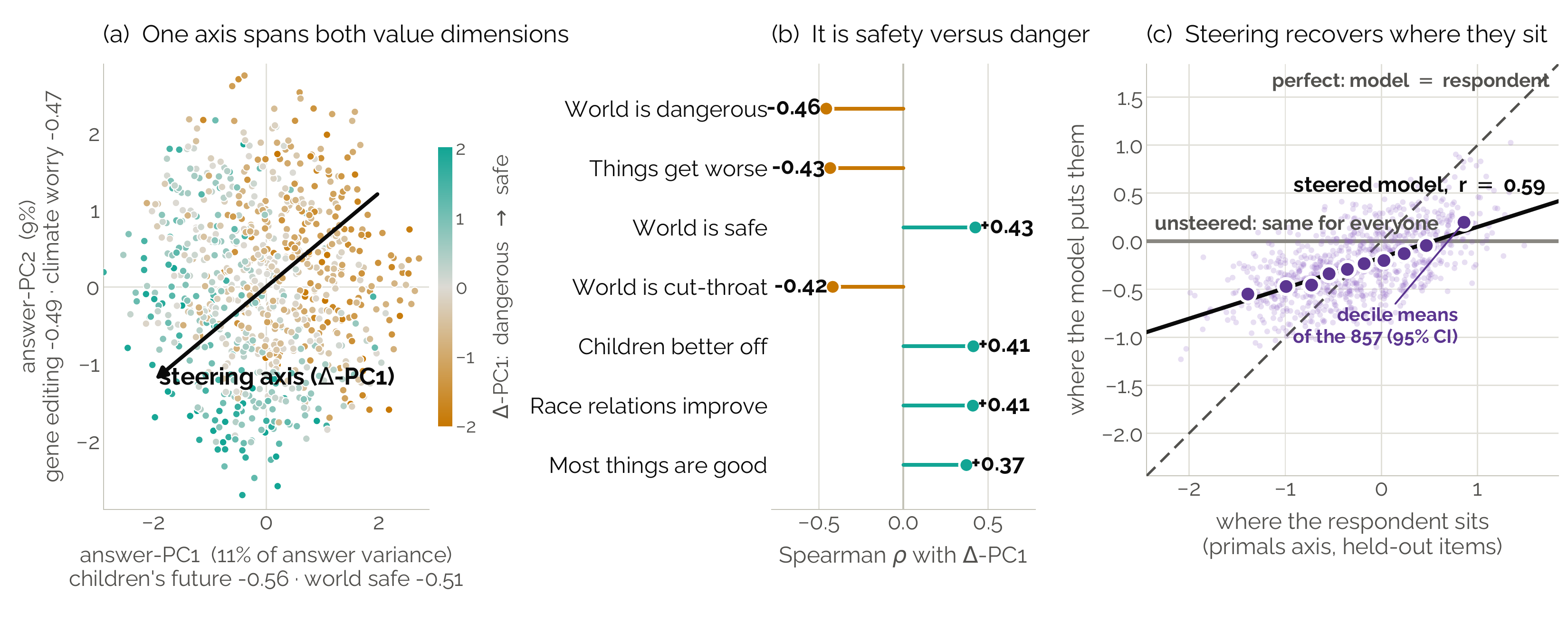}
    \caption{Interpretability results on PrimeX}
    \label{fig:primex-interpret}
\end{figure}

\begin{figure}[!t]
    \centering
    \includegraphics[width=0.7\textwidth]{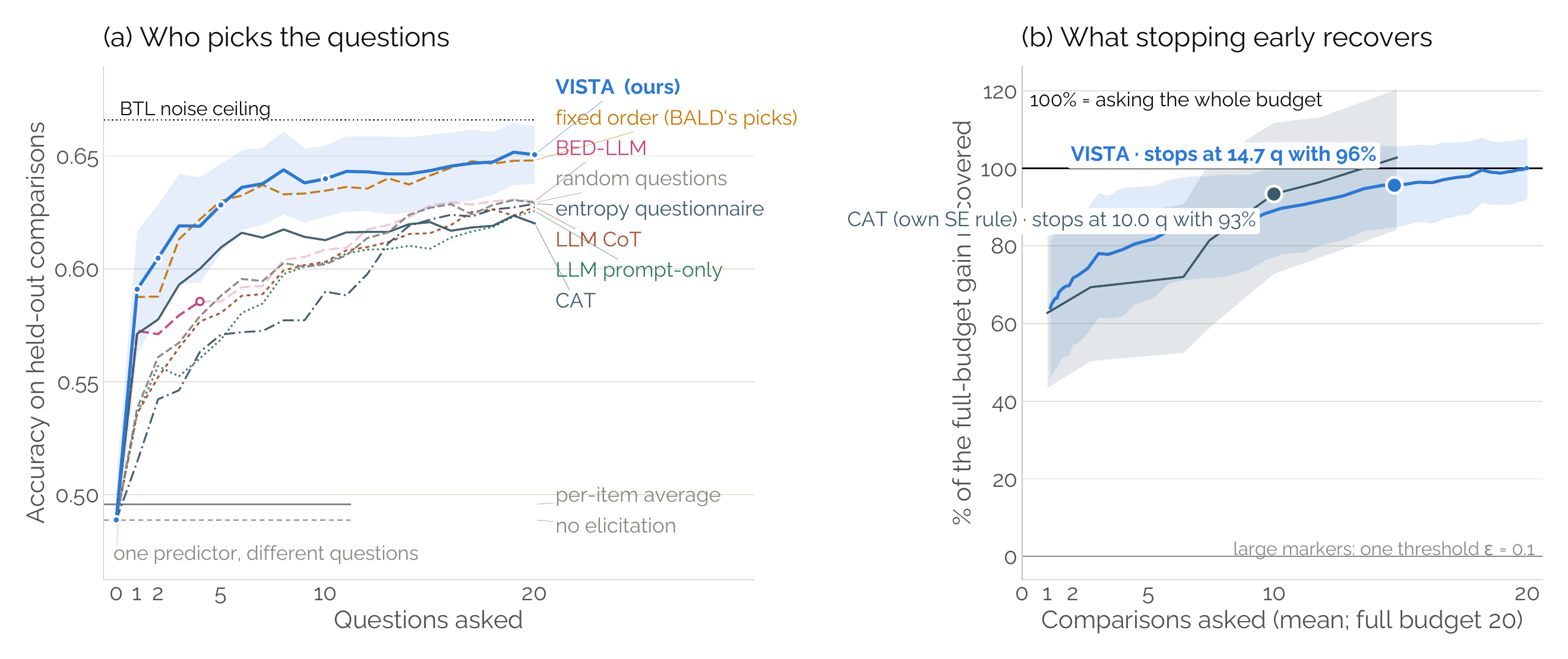}
    \caption{Elicitation results on food2d}
    \label{fig:elicit-food2d}
\end{figure}

\end{document}